%% file: paper.tex
\documentclass[preprint]{elsarticle}

\usepackage{lineno,hyperref}
\modulolinenumbers[5]

\journal{Applied Soft Computing}

\usepackage{tikz}
\usetikzlibrary{arrows,shapes,automata,backgrounds,petri,topaths,decorations.pathmorphing,3d,calc, positioning,intersections}	
\usepackage{amsmath}
\usepackage{array}
\usepackage{multirow,tabularx}
\usepackage{pdfpages}

\usepackage{amsmath}
\usepackage{multirow}
\usepackage{booktabs}
\usepackage{wasysym}
\usepackage[final]{todonotes}
\usepackage{paralist}
\definecolor{aoenglish}{rgb}{0.0, 0.5, 0.0}

\definecolor{responsecolor}{rgb}{0.0, 0.0, 0.0} % Changed to hide changes
\newcommand{\new}[1]{{\color{responsecolor}#1}}

\makeatletter
\def\tikz@lib@cuboid@get#1{\pgfkeysvalueof{/tikz/cuboid/#1}}
\input{cuboid-draw.tex}
\newcommand{\tikzcuboidset}{\@ifstar\tikzcuboidset@star\tikzcuboidset@nostar} 
\newcommand{\tikzcuboidset@nostar}[1]{\tikzcuboidreset\tikzset{cuboid,#1}}
\newcommand{\tikzcuboidset@star}[1]{\tikzset{cuboid,#1}}
\makeatother

\definecolor{colorgood}{HTML}{feef71}
\definecolor{colorbad}{HTML}{88669c}
\definecolor{colormeh}{HTML}{72acb5}
\pgfmathsetseed{2}
\definecolor{colorup}{HTML}{ff7f0e}
\definecolor{colordown}{HTML}{1f77b4}
\definecolor{colorupd}{HTML}{2ca02c}
\definecolor{colorela}{HTML}{c90000}
\definecolor{colorprop}{HTML}{ee9900}
\definecolor{colorclass}{HTML}{276dd7}
\tikzset{
	diagonal/.pic={
	\draw[colorup, thick] (-0.5, -0.05)
		\foreach \i in {1,...,10} {-- ++(0.1, 0.1*rnd)};
	\draw[colordown, thick] (-0.5, 0.3)
		\foreach \i in {1,...,10} {-- ++(0.1, -0.17*rnd)};
	\draw[colorupd, thick] (-0.5, -0.2)
		\foreach \i in {1,...,5} {-- ++(0.1, 0.2*rnd)}
		\foreach \i in {6,...,10} {-- ++(0.1, -0.15*rnd)};
	\draw[fill] (-0.15, 0.05) circle (0.08);
	\draw[thick] (-0.5, -0.5) rectangle (0.5, 0.5);
	}
}
\tikzset{
	highlevel/.pic={
		\draw[fill=colorgood, thick] (-0.5, 0) rectangle (0, 0.5);
		\draw[fill=colorbad, thick]  (0, 0.5) rectangle (0.5, 0);	
		\draw[fill=colormeh, thick]  (-0.5, -0.5) rectangle (0, 0);	
		\draw[fill=colorgood, thick] (0, -0.5) rectangle (0.5, 0);		
	}
}
\tikzset{
	similarity/.pic={
		\draw[fill=colorgood, thin] (0.35, -0.3) circle (0.1);
		\draw[fill=colorgood, thin] (0.30, -0.05) circle (0.1);
		\draw[fill=colorgood, thin] (0.10, -0.35) circle (0.1);
		\draw[fill=colormeh, thin] (-0.3, 0.25) circle (0.1);
	    \draw[thick] (-0.5, -0.5) rectangle (0.5, 0.5);
	}
}
\tikzset{
	searchspace/.pic={
		\draw[thin] (-2.5, -2.5) rectangle (2.5, 2.5);
		\draw[fill=black, thin] (-2, -1) circle (0.2);
		\draw[fill=black, thin] (-1, -2) circle (0.2);
		\draw[fill=black, thin] (0, 1) circle (0.2);
		\draw[fill=black, thin] (1, 2) circle (0.2);
		\draw[fill=black, thin] (2, -0) circle (0.2);
	}
}

\begin{document}

\begin{frontmatter}

%\title{Tools for Landscape Analysis of \\GAN-based Latent Vector Optimisation in Games}
\title{Tools for Landscape Analysis of Optimisation Problems in \new{Procedural Content Generation} for Games}

\author{Vanessa Volz}
\address{modl.ai, Denmark}
\cortext[mycorrespondingauthor]{Corresponding author}
\ead{vanessa@modl.ai}

\author{Boris Naujoks}
\address{TH Köln -- University of Applied Sciences, Germany}

\author{Pascal Kerschke}
\address{TU Dresden, Germany}

\author{Tea Tušar}
\address{Jožef Stefan Institute, Slovenia}

\begin{abstract}
The term Procedural Content Generation (PCG) refers to the (semi-)au\-to\-ma\-tic generation of game content by algorithmic means, and its methods are becoming increasingly popular in game-oriented research and industry. A special class of these methods, which is commonly known as search-based PCG, treats the given task as an optimisation problem. 
% with the aim of identifying the most suitable examples of valid content. 
% As such, the problem is based on a searchable representation of the space of generatable content as well as on a fitness function that can guide the optimisation algorithm's search. Further, it is usually treated as a black box and -- mainly due to the associated lack of knowledge about its structural properties -- 
Such problems are predominantly \new{tackled} by evolutionary algorithms.

We will demonstrate in this paper that % with the help of a popular Mario level generator, 
obtaining more information about the defined optimisation %problem can substantially improve our understanding of the generated content.
problem can substantially improve our understanding of how to approach the generation of content. 
To do so, we present and discuss three \new{efficient} analysis tools, namely diagonal walks, the estimation of high-level properties, as well as problem similarity measures. We discuss the purpose of each of the considered methods in the context of PCG and provide guidelines for the interpretation of the results received. This way we aim to provide methods for the comparison of PCG approaches and % of generated content and, 
eventually, increase %its quality.
the quality and practicality of generated content in industry.
\end{abstract}


\end{frontmatter}

%\linenumbers

\section{Introduction}

\begin{figure}[b]
    \centering
    \includegraphics[width=0.8\textwidth]{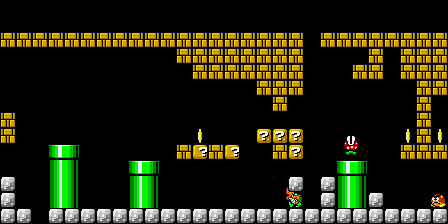}
    \caption{Exemplary (underground) level of Super Mario Bros.}
    \label{fig:marioLevel}
\end{figure}

Search-based procedural content generation is a very popular approach for generating various types of content for games, such as levels (for example for Super Mario Bros., see Figure~\ref{fig:marioLevel}) and weapons \cite{pcgbook2}. They work by formulating the generation process as an optimisation problem, where the task is to identify content that fulfils a given objective best.  
According to \cite{pcgbook2}, in order to define this optimisation problem, the following three components need to be specified:
 (1) a suitable \emph{representation} / search space for the content that can be searched (more details in Section~\ref{sec:setup:search}) 
 by (2) an (optimisation) \emph{algorithm}, which in turn is guided by (3) a suitable \emph{fitness function} (more details in Section~\ref{sec:setup:fitness}). 
 Most previous publications on search-based PCG focus on one aspect of the problem definition and treat the remaining components as given \cite{pcgbook12}, which limits our understanding of the problem as a whole. 

Of course, focusing only on one component allows more in-depth analyses and streamlined discussions. However, taking a more holistic and application-agnostic approach to analysing search-based PCG systems instead comes with several potential benefits. This means that all interconnections between the different components of a PCG system are accounted for in the analysis. For example, a better understanding of the fitness landscape, which is a product of representation and fitness function, can allow to select a suitable search algorithm \cite{kerschke2018survey,munoz2015_as,kerschke2018bbob,bischl2012,prager2020cc}. This selection is important as, according to the \emph{no free lunch} theorem \cite{wolpert1997}, no single algorithm can perform well on all types of problems. Aspects such as noise levels, number of local and global optima, or the landscape's ruggedness should be considered when selecting an algorithm \cite{malan2013,mersmann2011,munoz2015_as,kerschke2019flacco}. In addition, knowledge about global optima can help to assess the success of the PCG approach and the potential for \new{further optimisation}.

This information is not only helpful for the selection of an optimisation algorithm, but can also be utilised in order to choose the best representation and fitness function possible. For example, if the dimension of the representation is variable, information about the scalability of the PCG approach can help to determine the smallest \new{(i.e. lowest dimensional)} representation that still offers sufficient detail and variety. Usually, the smaller the representation, the easier it is to find fit solutions.
In addition, the content is usually intended to fulfil objectives around abstract game-related concepts, such as, e.g., game difficulty. These concepts, however, can usually be expressed and implemented in several ways, for example using different Artificial Intelligence (AI) game-playing agents for simulation.
These different implementations might not create fitness landscapes of the same type, and consequently influence the attainable content.

Finally, a more extensive analysis of search-based PCG applications from the standpoint of optimisation algorithms also produces information on the robustness of the proposed approach. In this context, robustness is especially important with regards to the reliability of a content generator to produce similar content and thus fulfil expectations, for example regarding its playability.
Further, different training examples and initialisation procedures could be tested to investigate the applicability of a given PCG algorithm to different games. This is especially true if all three components, i.e., representation, fitness function and search algorithm can be varied. Results from this type of analysis also facilitate comparisons between different content generators.

Research in evolutionary computation has been applying several methods of analysis for understanding and improving the behaviour of optimisation algorithms for several years \cite{cocoperf,bartz2020benchmarking}. In this paper, we
\begin{enumerate}
    \item present and discuss how one can apply several of these landscape analysis methods to (search-based) PCG;
    \item summarise how the gained insights can -- and should -- be used to evaluate and improve PCGs; and
    \item demonstrate the methods' applicability by means of an exemplary use case in which we analyse generated Mario levels. 
\end{enumerate}

In order to keep the paper concise, the methods we discuss in more detail in this paper are all intended for the analysis of fitness landscapes of black-box problems, independent of the applied optimisation algorithm.
This type of analysis is often a first step and especially relevant for PCG approaches due to the lack of suitable existing information helpful for selecting an optimiser.

We are further only describing the proposed methods in the context of continuous search spaces, since that is the type of problem we have chosen as an example application. All the required concepts also exist for optimisation problems with non-continuous search spaces, however, and are thus transferable.

Our contribution in this paper is thus a \textbf{vision and tutorial to study optimisation problems in search-based PCG from the perspective of systematic analysis tools}.
To support our vision, we further provide exemplary results for a benchmark called GBEA (Game Benchmark for Evolutionary Algorithms \cite{volz2019gbea}), which is based on PCG applications.
We are able to demonstrate that the analysis we propose is useful to strengthen the interpretability and thus usefulness of the aforementioned benchmark.

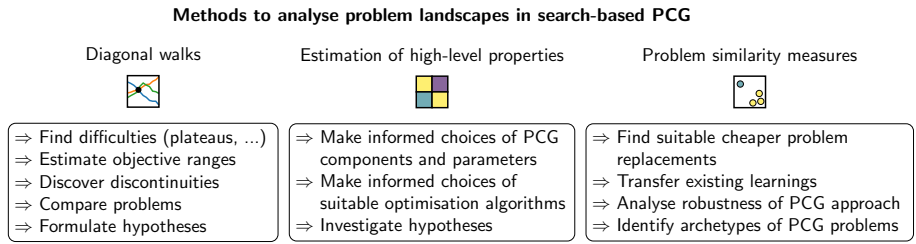
\begin{figure}[t]
    \resizebox{\textwidth}{!}{%
    \begin{tikzpicture}[
    	font=\sffamily\linespread{1}\selectfont, 
    	method/.style={rectangle, minimum width=35mm},
    	block/.style={rectangle, rounded corners, draw}
    ]
    	\draw (0, 0) node (m0) {\textbf{Methods to analyse problem landscapes in search-based PCG}};
    	\draw node[method, below=2mm of m0] (m2) {Estimation of high-level properties};
    	\draw pic[scale=0.6, below=4mm of m2] {highlevel};
    	\draw node[block, text width=52mm, below=10mm of m2] (t2)
    		{$\Rightarrow$ Make informed choices of PCG \\ \phantom{$\Rightarrow$ }components and parameters\\
    		 $\Rightarrow$ Make informed choices of \\ \phantom{$\Rightarrow$ }suitable \new{optimisation} algorithms\\
    		 $\Rightarrow$ Investigate hypotheses
    		};
    	\draw node[block, text width=49mm, left=2mm of t2] (t1)
    		{$\Rightarrow$ Find difficulties (plateaus, ...)\\
    		 $\Rightarrow$ Estimate objective ranges\\
    		 $\Rightarrow$ Discover discontinuities \\
    		 $\Rightarrow$ Compare problems\\
    		 $\Rightarrow$ Formulate hypotheses
    		};
    	\draw node[block, text width=60mm, right=2mm of t2] (t3)
    		{$\Rightarrow$ Find suitable cheaper problem \phantom{$\Rightarrow$ }replacements\\
    		 $\Rightarrow$ Transfer existing learning\new{s}\\
    		 $\Rightarrow$ Analyse robustness of PCG approach\\
    		 $\Rightarrow$ Identify archetypes of PCG problems		
    		};
    	\draw node[method, above=10mm of t1] (m1) {Diagonal walks};
    	\draw node[method, above=10mm of t3] (m3) {Problem similarity measures};
    	\draw pic[scale=0.6, above=6mm of t1] {diagonal};
    	\draw pic[scale=0.6, above=6mm of t3] {similarity};
    \end{tikzpicture}
    }
    \caption{Overview of the proposed methods and their purpose in the context of PCG. Diagonal walks are described in Section~\ref{sec:walks}, the estimation of structural high-level properties in Section~\ref{sec:heatMaps}, and problem similarity measures in Section~\ref{sec:tsne}.}
    \label{fig:scheme1}
\end{figure}

In the following, we first give some background information, starting with related work on landscape analysis as well as search-based PCG in Section~\ref{sec:rw}. We also specifically survey research that is intended to obtain more information on PCG approaches and identify a definite lack of such work. 
The general experimental setup for the experiments conducted in the following sections is provided in Section~\ref{sec:setup}.
In the second part of the paper, we propose several analysis tools and demonstrate their usefulness using an exemplary problem from the GBEA benchmark.
Tools that are discussed are diagonal walks in Section~\ref{sec:walks}, the estimation of structural high-level properties in Section~\ref{sec:heatMaps}, and problem similarity measures in Section~\ref{sec:tsne}. Each of these sections contains a short explanation of the \new{method in question}, followed by a description of its purpose in the context of PCG. An overview of the discussed tools and their purpose in PCG is given in Figure \ref{fig:scheme1}. We then demonstrate each tool's applicability \new{by applying it to} our example. Each section is concluded by a short discussion of the respective method's suitability, strengths and weaknesses.
We conclude the paper in Section~\ref{sec:final} with a summary of our concrete findings. Finally, we discuss our vision of using optimisation analysis tools to improve PCG research in the future.

% Random seed for the diagonal icon
\pgfmathsetseed{2}

% %% --------------------------------------------------------------------75
\section{Related Work}
\label{sec:rw}

In the following, we give an overview of related work. We start with an overview of Exploratory Landscape Analysis, which we use as an example of current approaches to understanding fitness landscapes in black-box optimisation problems. Next, we conduct a brief survey of the state-of-the-art of search-based PCG, focusing specifically on work that analyses the optimisation problems in PCG in more detail. 

\subsection{Exploratory Landscape Analysis}
\label{sec:ela}

Exploratory Landscape Analysis (ELA), sometimes also \new{called} fitness landscape analysis, 
stands for a sophisticated method that employs automatically computable (mainly numerical) values to extract representative information from a problem's landscape \cite{mersmann2011, munoz2015ic, kerschke2019flacco}. 
These numbers, the so-called \emph{features}, can be used to derive more tangible properties of the landscapes, such as whether they are rugged \cite{malan2009}, possess an underlying funnel structure \cite{kerschke2015}, or contain plateaus \cite{mersmann2011}. %Based on this information, features also
As they provide cheap and explicitly measurable surrogates for these (hardly quantifiable) high-level properties, automatically computable landscape features 
enable more sophisticated analyses on the problems at hand such as examining the underlying problem spaces \cite{Skvorc2020,munoz2019,lang2021exploratory,schneider2022hpo}, characterising algorithmic search behavior \cite{engelbrecht2021influence,Pitra2019} as well as selecting \new{and configuring} well-performing optimisation algorithms (\cite{bischl2012,munoz2015_as,kerschke2018bbob,Jankovic2020,prager2022automated} and \cite{belkhir2016,belkhir2017,prager2020cc}, respectively). Note that in order to save valuable resources, features are ideally computed on a very small set of sampled points \cite{kerschke2015,Renau2019}. 

Over the decades, a plethora of features have been proposed, which makes a detailed discussion in this work infeasible. Instead\new{,} we refer the interested readers to \cite{kerschke2019flacco,derbel2019,kerschke2018survey,kerschke2018bbob} for comprehensive overviews of recent works on this topic. 
For now, we only provide two features to give an \new{example} of typical features:
% (1) the coefficient of determination $R^2 \in [0, 1]$ (i.e., the quality) of a quadratic model that has been fitted to the sampled data \cite{mersmann2011}, and (2) 
% %the ratio between the average distance of the samples' points to their respective nearest neighbours and the average distance to their nearest \emph{better} neighbours (i.e., the nearest neighbour among all observations with a better fitness value) 
% \new{the ratio between (i) the average distance of the sample points to their respective nearest neighbour, and (ii) the average distance to their nearest \emph{better} neighbour (i.e., the nearest neighbour among all observations with a better fitness value)}
% \cite{kerschke2015}. 
\begin{enumerate}
    \item the coefficient of determination $R^2 \in [0, 1]$ (i.e., the quality) of a quadratic model that has been fitted to the sampled data \cite{mersmann2011}, and
    \item \new{the ratio between (i) the average distance of the sample points to their respective nearest neighbour, and (ii) the average distance to their nearest \emph{better} neighbour (i.e., the nearest neighbour among all observations with a better fitness value)} \cite{kerschke2015}.
\end{enumerate}
Both features are useful when trying to distinguish between rather unimodal functions ($R^2 \approx 1$) and random landscapes that rather look like an egg box ($R^2 \approx 0$). 

\subsection{Search-based Procedural Content Generation}
\label{sec:rw:pcg}
\label{sec:rw:pcg:land}

The (semi-)automatic generation of game content through algorithmic means is commonly referred to as procedural content generation (PCG) \cite{shaker2016procedural}. 
Various approaches to PCG exist, but an especially popular subset is dubbed search-based PCG \cite{searchPCG,pcgbook2}. In the corresponding approaches, some notion of fitness is required to evaluate generated content. 
The ``fittest'' content can then be discovered by exploring the content-representing search space, for which the fitness values are given by the fitness function. Most commonly, the search is performed by evolutionary algorithms, likely due to their ability to handle black-box fitness functions \cite{Yannakakis-A,pcgbook2}.

As in any optimisation problem, the difficulty of finding good content in search-based PCG depends on the structural challenges posed by the fitness landscape (i.e., the combination of representation and fitness function), as well as the search algorithm operating \new{on it}. However, previous work on the analysis of search-based PCG as optimisation problems \new{often} lacks a holistic approach.

For example, numerous publications address the difficulty of finding suitable representations for game content. Novel representations for different subsets of game content are proposed regularly (see \cite{pcgbook9} for an overview). One \new{often addressed challenge of representations} is the dimension of the search space \cite{searchPCG}, as the chosen representation is required to map to relatively complex content (\new{e.g.} a whole level). \new{Still}, if the dimension is reduced too much, the variety of valid content might be limited, resulting in less interesting gameplay. Other properties of representations that are often addressed are the locality \cite{searchPCG} and the scarcity of valid solutions \cite[pp.~93--95]{dagstuhl2017}. In \cite{Yannakakis-Experience}, finding good representations is identified as an open problem. Recent efforts have been targeting the creation of such rich search spaces for game content \cite{mariogan}\footnote{Brief video explanation: \url{https://www.youtube.com/watch?v=NObqDuPuk7Q}}. \new{Building on this, newer efforts have been aimed at transforming these search spaces to facilitate exploration of interesting parts of the search space \cite{manifold}.}

The problem of defining fitness functions, however, is usually addressed separately from the choice of representation. This is because finding an automatic evaluation function for generated content without feedback from human players is an ill-posed problem, as subjective human perception needs to be expressed and formalised \cite{searchPCG}. For this reason, numerous fitness functions have been proposed in literature, especially for heavily researched games, such as board- \cite{Browne-Automatic} and platform games \cite{marioEval}. Several (largely agreeing) attempts have also been made in order to characterise these evaluation functions \cite[pp.~122--125]{dagstuhl2017}, \cite{Yannakakis-Experience, searchPCG, pcgbook2}. 

Existing evaluation approaches of both components (fitness function and representation) largely revolve around exploratory assessment of generated content, especially by visualising its variety -- for example via expressive range analysis \cite{pcgbook12}. 
The improvement of the search algorithm's fitness over iterations is usually reported as well, but this only offers limited interpretability if the fitness landscape is unknown.

It is therefore indispensable to analyse the fitness landscapes of search-based PCG. However, related work in this area is sparse. There are a few publications where search spaces are characterised based on the performance of search algorithms. For example, in \cite{landscapeautomata}, several versions of an evolutionary algorithm are used to optimise landscape automata based on two fitness functions. The study, however, focuses on identifying good parameters for the evolutionary algorithms and for the representation, instead of executing a landscape analysis. Nevertheless, this type of analysis allowed the authors to characterise the landscape as multimodal, i.e., containing multiple local and/or global optima \cite{preuss2015}. Another  publication \cite{hearthstone} uses a similar approach in order to describe the available gameplay to a beginner in a popular trading card game. The fitness function of evolving decks was shown to be jagged (i.e., it showed small local irregularities) and in their experiments, \new{but the results are mainly anecdotal and tied to the optimisation approach}

The only formal landscape analysis of search-based PCG we were able to find is based on a specific representation for maze levels, called apoptotic cellular automata \cite{6463449}. The study is based on a survey of fitness landscape analysis methods \cite{Pitzer2012}, which were adapted appropriately for application in this context. The authors find that the fitness landscape is rugose, i.e., it possesses a multitude of local optima and large plateaus with low fitness. Based on their analysis, the authors see a similarity to Shekel's foxhole function \cite{shekel}.

Unfortunately, the results are difficult to generalise. The problem in the study uses a specific, non-standard representation, while in search-based PCG, continuous representations are usually preferred \cite{searchPCG}. In addition, only a single fitness function is investigated. As evidenced by taxonomies and surveys \cite{Yannakakis-Experience}, there is a variety of fitness functions, which will likely produce very different landscapes. 

In this paper, we extend previous work in several ways. We analyse a problem with a more common representation, in conjunction with several (28) fitness functions.

Some literature can also be found on analysing the landscapes of games from the perspective of a game-playing agent. A recent example of this type of work is \cite{landscapeRL}\new{,} where various games are characterised based on graph representations of playthroughs. The authors propose to use the resulting representation to be able to compare different games and to facilitate the understanding of game agents. In contrast, in this paper, we compare different problems in search-based PCG to facilitate the understanding of PCG algorithms. The results and representations are thus unfortunately not transferable.

% %% --------------------------------------------------------------------75
\section{Example Application MarioGAN}
\label{sec:setup}

Super Mario Bros., or Mario for short, is a classic platform game (``platformer'') and has been targeted by various research efforts, including competitions and publications on game AI \cite{togelius2013championship}, as well as procedural generation of levels \cite{Horn-A}. A survey of several fitness functions that have been used in previous research can be found in \cite{marioEval}. 

As is clear from several surveys, a multitude of level generators for Mario \cite{Horn-A} and similar platformers exist. However, as described in the \new{previous} section, the search landscapes of the corresponding generators have not been analysed in detail. Several publications address different aspects of the problem, such as fitness functions \cite{marioEval} or search space \cite{mariogan}. As we are not aware of any holistic analyses of optimisation problems related to Mario level generation, we are using this application as an example to demonstrate the methodology we propose in this paper.  We thus select a previously published search-based PCG approach (dubbed MarioGAN \cite{mariogan}) that has certain \new{representative} characteristics, i.e., the representation \new{is} a continuous vector \cite{searchPCG} and suitable for evolutionary algorithms \cite{pcgbook2, Yannakakis-A}.

\subsection{Search Space}
\label{sec:setup:search}
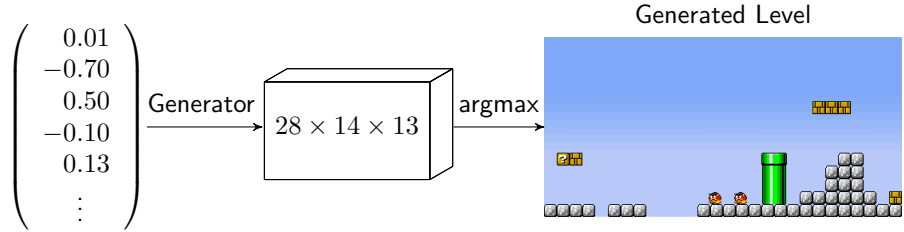
\begin{figure}
\tikzstyle{topic} = [rectangle, inner sep=0pt]
\centering
\begin{tikzpicture}[node distance=1.7cm,>=stealth',bend angle=45,auto, font=\sffamily]  
\tikzstyle{every label}=[font=\sffamily\small, label distance=0.3cm, text=gray] 		

\begin{scope}
\node [topic] (t1) {$\left( \begin{array}{r} 0.01 \\ -0.70 \\ 0.50 \\ -0.10 \\ 0.13 \\  \vdots\phantom{00} \end{array} \right)$};
\node [topic] (t2) [right of=t1, xshift=2.5cm]{};
\node [topic] (t3) [right of=t2, align=center, xshift=2.7cm] {\includegraphics[scale=0.3]{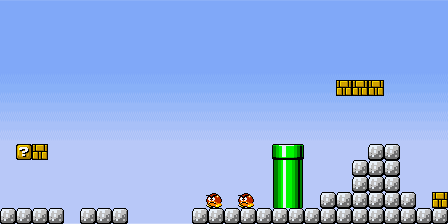}};
\end{scope}
\pic[black] at (2.5,-0.7,0) {cuboid=2.2--1.3--0.7};
\path[->] (t1)  edge node {Generator\phantom{j}} ([xshift=-1.7cm]t2.west);
\path[->] ([xshift=0.79cm]t2.east)  edge node {argmax} (t3);
\node[topic] (tt1) [above of=t1, yshift=0.6cm] {Latent Vector};
\node[topic] (tt2) [right of=t1, xshift=1.9cm] {$28 \times 14 \times 13$};
\node[topic] (tt3) [above of=t3, yshift=-0.2cm]{Generated Level};
\end{tikzpicture}
\caption{A schematic representation of the level generation process using MarioGAN.}
\label{fig:mariogan}
\end{figure}

For our analysis, we use the level generator proposed in \cite{mariogan}, which we will call MarioGAN in the following. The generator is a neural network that takes an input vector \new{with values between $-1$ and $1$} and produces $13$ matrices of size $28 \times 14$. The search space is thus \new{bounded and} continuous even though the representation of the levels is discrete. The size of the input vector just depends on the structure of the neural network and can be chosen arbitrarily. These are translated to $28 \times 14$-dimensional Mario levels using a binary encoding of $13$ different tile types. A visualisation can be found in Figure~\ref{fig:mariogan}.
The generator is trained using an adversarial approach (Generative Adversarial Networks - GANs) as suggested in \cite{goodfellow2014generative,arjovsky2017wasserstein} and is trained on Super Mario Bros.\ original levels contained in the video game level corpus \cite{Summerville:pcg2016-VGLC}.

Using this approach, generators for different input space dimensions can be trained. This allows us to analyse the scaling behaviour of the executed search algorithms.
Furthermore, the levels are generated within milliseconds, resulting in practical experiments in terms of computational resources.

\subsection{Fitness Functions}
\label{sec:setup:fitness}
For our analys\new{i}s, we are using the fitness functions proposed in the game-benchmark for evolutionary algorithms described in \cite{volz2019gbea}. These functions are based on the state of the art of platformer evaluation \cite{marioEval} and selected to produce diverse fitness landscapes. A short description of the measures computed for the functions along with their original sources is found below and a more formal definition with more details on how they are transformed into minimisation problems can be found in
~\ref{sec:appendix}:
\begin{compactitem}
    \item[\bf enemyDistribution:] 
    Horizontal distribution of enemies across the level. 
    The more enemies are grouped together, the more difficult it is to evade them. Measured as standard deviation of the enemies' x-axis coordinates \cite{marioEval}.
	\item[\bf positionDistribution:] 
	Vertical distribution of platforms across the level. 
	Platforms at various different heights typically seem interesting to a player. Measured as standard deviation of y-axis coordinates of tiles one can stand on \cite{marioEval}.
	\item[\bf decorationFrequency:] 
    Amount of non-standard tiles in the level, i.e., tiles that make the level interesting. 
	Measured as fraction of \emph{pretty tiles} := $\{$Tube, Enemy, Destructible Block, Question Mark Block, or Bullet Bill Shooter Column$\}$ \cite{marioEval}.
	\item[\bf negativeSpace:] 
    Amount of empty space in the level. 
	This characterises the way the player moves through the level. Measured as the fraction of tiles one cannot stand on \cite{Canossa2015TowardsAP}.
	\item[\bf leniency:] 
    A quantification of how easy it is to pass the level. 
	Measured as weighted sum of subjective \emph{leniency} of tiles as defined in \cite{mario100}.
	\item[\bf basicFitness:] 
    Difficulty for an AI to pass the level (according to the score used in the MarioAI championships \cite{togelius2013championship}). 
	Measured as a linear combination of several performance-related aspects, such as the distance of the level that was covered and the amount of collected coins.
	\item[\bf airTime:] Describes how much the AI agent jumped to pass through the level. Jumping is the main mechanic in Mario, so a level should require a considerable amount of it. Measured as the ratio between time in the air divided by time spent on the level. If the level is not completed, a penalty value is returned instead \cite{mariogan}.
	\item[\bf timeTaken:] 
    Time required by the AI to navigate through the level -- longer times likely involve some backtracking or other challenging parts. 
	Measured as the ratio between time taken and total time allowed for the level. If the level is not completed, a penalty value is returned instead \cite{mariogan}.
\end{compactitem}

\subsection{Optimisation Problems}
\label{sec:setup:prob}
\begin{figure}
    \centering
    \includegraphics[width=0.48\textwidth]{Pics/LevelClipped3.png}
    \includegraphics[width=0.48\textwidth]{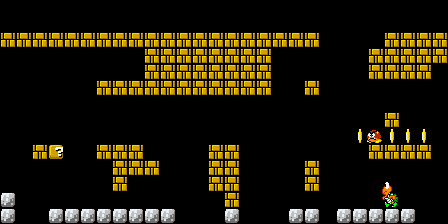}
    \caption{Exemplary overworld (left) and underground (right) levels.}
    \label{fig:overunder}
\end{figure}

Our set of MarioGAN optimisation problems is defined by combining the search space (Section~\ref{sec:setup:search}) with the fitness functions (Section~\ref{sec:setup:fitness}). Note that\new{,} without loss of generality\new{,} all problems have been turned into minimisation problems. In addition, several variations of each problem are provided to investigate different questions, resulting in a total of 28 problems listed in Table~\ref{mario:over}. Only variations resulting in functions with interesting and diverse fitness landscapes are included in the benchmark. These variations are:
\begin{compactitem}
    \item AI: Two different AIs (Baumgarten's A* and Scared \cite{togelius2013championship}) are implemented to simulate player behaviour for functions that require it.
    \item Training levels: GANs are trained separately on two disjoint sets of Mario levels, overworld (o) and underground (u) (see the images in Figure~\ref{fig:overunder}).
    \item Concatenation: Multiple level segments generated by a GAN can be concatenated (c) to create a longer level.
\end{compactitem}

\emph{Instances} of each problem (as defined in \cite{coco}) are created by varying the random seed used to initialise the training of the generators. In \new{this paper} we consider seven instances for \new{each} MarioGAN problem.

\begin{table}[!tb]
\centering
\caption{\label{mario:over}Overview and characterisation of the 28 fitness functions in the Mario suite ($m_1 - m_{28}$). A* refers to % by Robin Baumgarten, 
the winner of the first MarioAI gameplay competition \protect\cite{togelius2013championship}, which is based on an A* algorithm. Scared refers to \emph{ScaredAgent}, one of the default agents within the MarioAI competition framework, which tries to avoid all obstacles and enemies by jumping. Letters [o] and [u] specify the training levels used and [c] whether level segments need to be concatenated.}

\renewcommand{\tabcolsep}{10pt}
\renewcommand{\arraystretch}{0.65}
\begin{tabular}{llllll}
\toprule
\bfseries Fitness Measure & \bfseries AI  
& \bfseries o & \bfseries u & \bfseries oc & \bfseries uc\\
\midrule
enemyDistribution &  -   & $m_1$ & $m_2$ & &\\ \cmidrule{2-6}
positionDistribution &  -    & $m_3$ & $m_4$ & &\\ \cmidrule{2-6}
decorationFrequency &  -    & $m_5$ & $m_6$ & &\\ \cmidrule{2-6}
negativeSpace & -   & $m_7$ & $m_8$ & &\\ \cmidrule{2-6}
leniency & -    & $m_9$ & $m_{10}$ & &\\ \cmidrule{2-6}
\multirow{2}{*}{basicFitness} &  A*   & $m_{11}$ & $m_{12}$ & $m_{13}$ & $m_{14}$\\
  & Scared   & $m_{15}$ & $m_{16}$ & &\\ \cmidrule{2-6}
\multirow{2}{*}{airTime} &  A*  & $m_{17}$ & $m_{18}$ & $m_{19}$ & $m_{20}$\\
 &  Scared  & $m_{21}$& $m_{22}$ & &\\ \cmidrule{2-6}
\multirow{2}{*}{timeTaken} &  A*  & $m_{23}$ & $m_{24}$ & $m_{25}$ & $m_{26}$\\
 & Scared  &$m_{27}$ & $m_{28}$ & &\\ \bottomrule
\end{tabular}

\vspace*{-0.25cm}
\end{table}

% %% --------------------------------------------------------------------75
\section{Diagonal Walks}
\label{sec:walks}

\begin{figure}[t]
    \begin{tabular}{@{}c@{}|@{}c@{}}
			\begin{tabular}{@{}l@{}}
				% Random seed for the diagonal icon
				\pgfmathsetseed{2}%
				% Diagonal walks
				\resizebox{0.5\textwidth}{!}{%
				\begin{tikzpicture}[font=\sffamily\linespread{1}\selectfont]%
					\draw node[text width=85mm] (p0) {%
					For each GBEA problem $f$\new{, repeat a number of times}:\\
					\phantom{0}1. Choose a random anchor point in the search space\\
					\phantom{0}2. Repeat a number of times:\\
					\phantom{02. }2.1 Choose random direction\\
					\phantom{02. }2.2 Evaluate equidistant points along the straight \phantom{02. 2.2 }line defined by the anchor point and the chosen \phantom{02. 2.2 }direction};
					\draw node[above=4mm of p0.north west, anchor=west, xshift=8mm] (tit) {\textbf{Diagonal walks}};
					\draw pic[scale=0.6, left=4mm of tit] {diagonal};
				\end{tikzpicture}}
				
				\\\hline
				\\[-12pt]
				% Estimation of high-level properties
				\resizebox{0.5\textwidth}{!}{%
				\begin{tikzpicture}[font=\sffamily\linespread{1}\selectfont]
					\draw node[text width=85mm] (p0) {%
					For each high-level property $p$:\\
					\phantom{0}1. Train \textcolor{colorclass}{classifier(s)} on BBOB problems using\\ \phantom{02. }\textcolor{colorela}{ELA features}\\
				% 	%\phantom{0}2. Select the best one\\
					\phantom{0}2. Use \textcolor{colorclass}{classifier(s)} to predict property $p$ based on\\ \phantom{02. }\textcolor{colorela}{ELA features} of GBEA problems};
					\draw node[above=4mm of p0.north west, anchor=west, xshift=8mm] (tit) {\textbf{Estimation of high-level properties}};
					\draw pic[scale=0.6, left=4mm of tit] {highlevel};
				\end{tikzpicture}}
				
				\\\hline
				\\[-12pt]
				% Problem similarity measures
				\resizebox{0.5\textwidth}{!}{%
				\begin{tikzpicture}[font=\sffamily\linespread{1}\selectfont]
					\draw node[text width=85mm] (p0) {%
					1. Compute \textcolor{colorela}{ELA features} on GBEA problems\\
					2. Compute \textcolor{colorela}{ELA features} on other problems (optional)\\
					3. Use t-SNE on normalized \textcolor{colorela}{ELA features} to visualize\\ \phantom{0. }problems in 2-D};
					\draw node[above=4mm of p0.north west, anchor=west, xshift=8mm] (tit) {\textbf{Problem similarity measures}};
					\draw pic[scale=0.6, left=4mm of tit] {similarity};
				\end{tikzpicture}}
			\end{tabular}
			&
			\begin{tabular}{@{}c@{}}
			
			% Compute ELA features for problem f
				\resizebox{0.48\textwidth}{!}{%
				\begin{tikzpicture}[
					font=\sffamily\linespread{1}\selectfont, 
					block/.style={rectangle, rounded corners, draw, text centered, minimum height=10mm}
				]
					\draw node[block, minimum width=40mm, minimum height=12mm, text depth=12mm] (lhs) {Latin hypercube sampling};
					\draw pic[scale=0.2, yshift=-10mm] at (lhs) {searchspace};
					\draw node[yshift=-5mm] at (lhs) {$x_i$};
					\draw node[right=5mm of lhs.center, yshift=-4mm, text width=10mm, font=\sffamily\small\linespread{0.8}\selectfont] {search space};
					\draw node[block, text width=30mm, below=5mm of lhs] (eval) {Solution evaluation\\[3pt]$f(x_i)$};%\\[3pt]for $i = 1, \dots, 50D$};
					\draw node[block, minimum width=15mm, right=of $(lhs.east)!0.5!(eval.east)$] (f) {\texttt{flacco}};
					\draw node[block, text width=15mm, right=of f, fill=colorela, fill opacity=0.2, text opacity=1, draw=colorela, thick] (ela) {97 ELA features};
					\draw [-stealth] (lhs) -- (eval);
					\draw [-stealth] (lhs) -- (f);
					\draw [-stealth] (eval) -- (f);
					\draw [-stealth] (f) -- (ela);
					\coordinate (C) at ($(lhs.west)!0.5!(ela.east)$);
					\path let \p1 = (C), \p2 = (lhs.north) in node[yshift=5mm] at (\x1, \y2) {\textbf{Compute ELA features for problem $f$}};
					\draw [white] let \p1 = (C), \p2 = (lhs.north) in ([yshift=1mm]\x1, \y2) -- ++(-48mm, 0) -- ++(96mm, 0);
				\end{tikzpicture}}
				\\\hline
				\\[-11pt]
				% Train a classifier
				\resizebox{0.48\textwidth}{!}{%
				\begin{tikzpicture}[
					font=\sffamily\linespread{1}\selectfont, 
					block/.style={rectangle, text centered, rounded corners, draw, minimum height=10mm}
				]
					\draw node[block, text width=30mm, draw=colorela, thick] (bb) {ELA features for BBOB problems};
					\draw node[block, text width=30mm, below=5mm of bb] (ann) {Hand-annotated property $p$ for BBOB problems};
					\draw node[block, text width=15mm, right=of $(bb.east)!0.5!(ann.east)$] (ml) {ML algorithm};
					\draw node[block, text width=15mm, below=5mm of ml] (sel) {Feature selection};
					\draw node[block, text width=15mm, right=of ml, fill=colorclass, fill opacity=0.2, text opacity=1, draw=colorclass, thick] (class) {Classifier};
					\draw [-stealth] (bb) -- (ml);
					\draw [-stealth] (ann) -- (ml);
					\draw [stealth-stealth] (sel) -- (ml);
					\draw [-stealth] (ml) -- (class);
					\coordinate (C) at ($(bb.west)!0.5!(class.east)$);
					\path let \p1 = (C), \p2 = (bb.north) in node[yshift=5mm] at (\x1, \y2) {\textbf{Train a classifier for high-level property $p$}};
					\draw [white] let \p1 = (C), \p2 = (bb.north) in ([yshift=1mm]\x1, \y2) -- ++(-48mm, 0) -- ++(96mm, 0);
				\end{tikzpicture}}
				\\\hline
				\\[-11pt]
				% Predict with classifier
				\resizebox{0.48\textwidth}{!}{%
				\begin{tikzpicture}[
					font=\sffamily\linespread{1}\selectfont, 
					block/.style={rectangle, text centered, rounded corners, draw, minimum height=10mm}
				]
					\draw node[block, text width=22mm, draw=colorela, thick] (ela) {ELA features for problem $f$};
					\draw node[block, text width=15mm, right=of ela, draw=colorclass, thick] (class) {Classifier};
					\draw node[block, text width=18mm, right=of class] (res) {Property $p$};
					\draw [-stealth] (ela) -- (class);
					\draw [-stealth] (class) -- (res);
					\coordinate (C) at ($(ela.west)!0.5!(res.east)$);
					\path let \p1 = (C), \p2 = (ela.north) in node[yshift=5mm] at (\x1, \y2) {\textbf{Predict high-level property $p$ for problem $f$}};
					\draw [white] let \p1 = (C), \p2 = (ela.north) in ([yshift=1mm]\x1, \y2) -- ++(-48mm, 0) -- ++(96mm, 0);
				\end{tikzpicture}}
			\end{tabular}
    \end{tabular}
    \caption{Pseudocode for diagonal walks, estimation of high-level properties and problem similarity measures on the left hand side with additional explanation of the nontrivial steps on the right hand side. The derivation of ELA features (red) and classifiers (blue) is highlighted in the right hand part as well as their \new{utilisation} in the whole figure.}
    \label{fig:scheme2}
\end{figure}
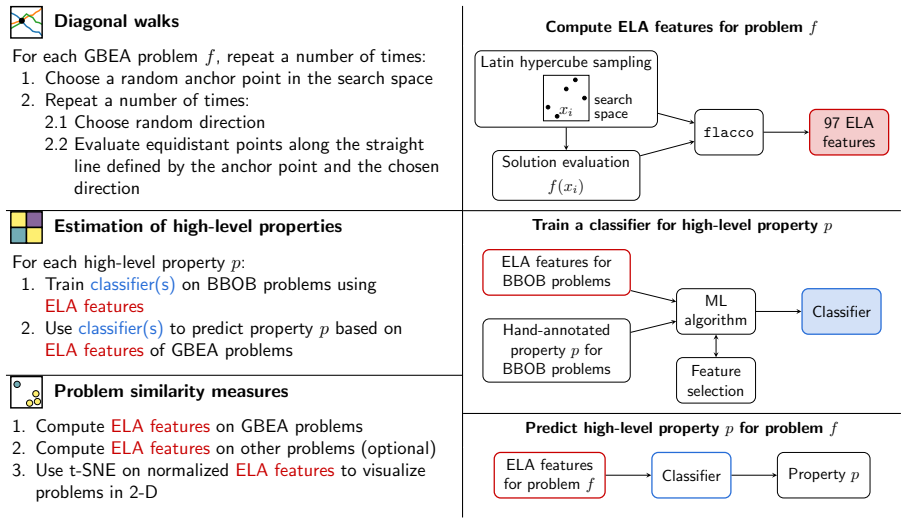

In the following, as well as in the two upcoming sections, different methods for analysing PCG approaches are discussed in detail (they are also presented in Figure~\ref{fig:scheme2}). We start with the most basic method, diagonal walks,
which are a simple tool for a first analysis of optimisation problems.

% %% --------------------------------------------------------------------75
\subsection{Method Description}
\label{sec:ana:walk}

According to \cite{Pitzer2012,munoz2015ic}, landscape walks are a useful tool for landscape analysis. Following the example from \cite{volz2019gbea}, we perform a type of landscape walks called diagonal walks through a random point. We generate a random anchor point that represents a valid solution and \new{a random vector representing a direction. Together, they define a line through the search space that is limited by the search space boundaries. Then, we ``walk'' on this line from one end to the other in equidistant steps passing through the anchor point.} 
As the randomly chosen directions are generally not axis-aligned, this means that all variable values are changed concurrently. 
Because the position and the direction of such walks are random, the number of available steps differs from walk to \new{walk, the anchor point is not necessarily at the center of the walk and the starting and ending points of the walk are not necessarily on the search space boundary. These walks are repeated several times by using different directions for the same anchor point in order to put the different walks into context of each other. Furthermore, to increase coverage, this procedure should be repeated with various anchor points.}

\begin{figure*}[t]
\centering
\begin{tabular}{cc}
\includegraphics[height=100pt,trim={0.2cm 5pt 38.1cm 35pt},clip, page=1]{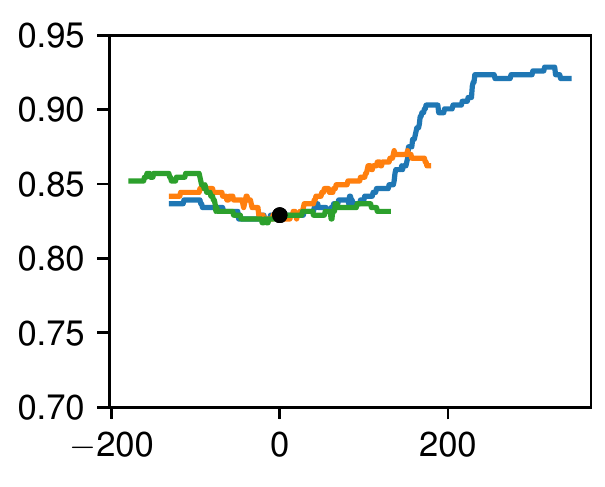} &
\includegraphics[height=100pt,trim={0.2cm 5pt 38.1cm 35pt},clip, page=2]{Pics/rw-gan-mario-diagonal-for-paper-selected.pdf} \\
\footnotesize{(a) negativeSpace overworld ($m_7$)} & 
\footnotesize{(b) negativeSpace underground ($m_8$)} \\
\includegraphics[height=100pt,trim={0.2cm 5pt 38.1cm 35pt},clip, page=3]{Pics/rw-gan-mario-diagonal-for-paper-selected.pdf} &
\includegraphics[height=100pt,trim={0.2cm 5pt 38.1cm 35pt},clip, page=4]{Pics/rw-gan-mario-diagonal-for-paper-selected.pdf} \\
\footnotesize{(c) basicFitness A* overworld ($m_{11}$)} & 
\footnotesize{(d) basicFitness Scared overworld ($m_{15}$)} \\ 
\end{tabular}
\caption{\label{fig:walk-mario} Diagonal walks for (the first instance of) different fitness functions. Their problem IDs (see Table~\ref{mario:over}) are given in brackets. The different colours indicate separate walks through the same anchor point (black dot, centered at $x=0$). The $x$-axes show the number of steps taken in each direction from the anchor point, whereas the $y$-axes show the respective fitness values.}
\end{figure*}

\new{In our example, we perform diagonal walks with a single anchor point and three different directions.} 
In their visualisations (see Figure~\ref{fig:walk-mario}), 
the $x$-axis shows the number of steps along the diagonal line (with the anchor point being positioned at $x = 0$), and the $y$-axis depicts the corresponding function values. To enable a direct comparison of different optimisation problems, we use the same anchor point and directions throughout our analysis in this paper.

%

% %% --------------------------------------------------------------------75
\subsection{Purpose in Context of PCG}

As complex game content must be presented in a way that is comprehensible for evolutionary algorithms, the genotype-phenotype mappings used in the context of search-based PCGs are often very complex. In addition, the selected genotype might require the definition of non-standard variation operators in order to ensure that all generated content is valid.

For this reason, it is helpful to gain insights into how an algorithm explores the space of the generatable content. Diagonal walks are \new{an effective} tool to achieve this purpose. While they certainly do not provide a holistic picture, these walks can still serve as early indicators of difficulties that the search algorithm may face, such as the existence of plateaus or many narrow basins of attraction (multi-modality). If such issues are discovered, different options for representation and variation operators can be explored. Otherwise, \new{a suitable} optimisation approach can be chosen \new{by considering} the expected difficulties. For example, restarts are a common approach for handling multi-modal landscapes.

Furthermore, it is usually difficult to judge what is a comparatively good fitness value. Diagonal walks are a useful tool for a rough approximation of the achieved value ranges. Also helpful in this regards are statistics on the distribution of fitness values from random samples of the search space. Diagonal walks, however, give an impression of the relative positions of the sample solutions in the search area. 
This can help to identify whether the fitness landscapes contain discontinuities in objective space or whole areas with good objective values.

An additional important benefit of diagonal walks comes from the fact that many approaches rely on simulation-based measures computed during play\-throughs with AI players \cite[\ref{sec:appendix}]{Volz19}. Thus, comparing the fitness values computed from different AI simulations at the same points (i.e., same generated content) can give first insights into differences and similarities of player behaviours. This analysis is especially meaningful if diagonal walks from AI playthroughs are compared with corresponding human ones.

% %% --------------------------------------------------------------------75
\subsection{MarioGAN Results}
\label{sec:walks:gbea}

\subsubsection{Analysis}

The diagonal walks for some representative problems in MarioGAN can be found in Figure~\ref{fig:walk-mario}. Several interesting observations can be made \new{based on} these plots.
For example, the genotype-phenotype mapping as described in Section~\ref{sec:setup:search} maps a continuous latent space to a discrete level. Steps in the search space thus translate to the addition, removal, or swapping of one tile in a level to another. If a fitness function (such as negativeSpace) is computed directly on the level encoding, we would thus expect a step-like landscape with small steps indicating when tiles change. Exactly this behaviour can be observed in Figure~\ref{fig:walk-mario} (a) and \ref{fig:walk-mario} (b). Although the size of the steps differs for different instances and areas of the search space, it is present for all problems with representation-based fitness functions ($m_1$-$m_{10}$).

Despite being based on the same genotype-phenotype mapping, steps can not be observed when using simulation-based fitness functions (problems $m_{11}$-$m_{28}$). This is likely because changing a single tile in the level does not necessarily result in different agent behaviour. As such, the fitness measure basicFitness (i.e., the performance of the AI) as depicted in Figure~\ref{fig:walk-mario} (c) and (d), for example, does not necessarily change if a platform is added that can not be reached by Mario. However, the addition of a single enemy can significantly affect player behaviour and thus \new{the resulting} score.

The size of these effects also depends on the agent used for simulations. This can be seen when comparing the scores that A* and the ScaredAgent received for the same levels as shown in Figure~\ref{fig:walk-mario} (c) and (d), respectively. The naive ScaredAgent is more sensitive to changes, producing a widely varying performance. While the basicFitness still varies significantly for the A* agent overall, the variation per step is much smaller. Still, the lack of smoothness in both functions should be a large concern when picking an optimisation algorithm. Problems \new{like the ones discussed above} with a low locality are difficult for standard evolutionary algorithms \cite{rothlauf}, \new{for example}. %\pk{shouldn't we support this claim with a reference?}

Another observation is that the achieved fitness values for negativeSpace tend to be lower for underground levels ($m_8$) than overworld levels ($m_7$). This is expected, as underground levels \new{mimic} tunnels in dungeons and \new{are} therefore capped by ceilings. We have visually verified that this is indeed the case for the generated levels (see Figure~\ref{fig:overunder}). Through further \new{experiments} -- for which a detailed description would exceed the scope of this paper -- we were also able to demonstrate \new{the observed tendency} empirically.

\subsubsection{Interpretation}
\label{sec:walks:interpretation}

Based on the above analysis, it seems that evolutionary algorithms are well suited for optimising problems $m_1$-$m_{10}$ with representation-based fitness functions, as locality assumptions are fulfilled. This seems to indicate that the chosen genotype-phenotype mapping is suitable for the generated content.

However, problems with simulation-based fitness functions show landscapes with large local variations. This might require the development and application of techniques that are suited to handle such variations appropriately. In addition, the cause for these large variations should be investigated in more detail. One possible explanation for the variations is the noise in the fitness evaluations that is not taken into account, caused either by the AI or the physics engine. 

The large differences between fitness landscapes for different AI players also demonstrate how sensitive content evaluations react to the actual AI implementation used for the simulation. This illustrates the potential pitfalls of evaluating games content \new{exclusively} automatically, as discussed in \cite{volzCog2019}. Instead, it suggests that experiments using simulation-based fitness functions should at least be repeated with different AIs. Ideally, these AI \new{agents} resemble different player types in order to ensure diverse and meaningful behaviour \cite{personas}.

\subsection{Concluding Remarks}

Diagonal walks enable an investigation of how small changes in search space are reflected in objective space. Resulting observations of course only correspond to the selected random point and directions, and cannot offer \new{reliable} insights \new{into} global problem properties.

Still, this approach offers \new{an efficient} way to (visually) inspect characteristics of a high-dimensional space. Diagonal walks are easy to generate and interpret without the need for in-depth knowledge of evolutionary algorithms. They can thus serve as a standard method for sanity checking of PCG approaches. Since the number of samples required for this type of analysis is small, it is suitable for the often expensive fitness evaluations required for PCG. Furthermore, diagonal walks and their visualisation can serve as inspiration for the definition of hypotheses about fitness landscapes that can be tested with more sophisticated methods.

% %% --------------------------------------------------------------------75
\section{Estimation of High-Level Properties}
\label{sec:heatMaps}

If aimed at the general global structure of a fitness landscape, the aforementioned hypotheses can usually be formulated using high-level properties such as multi-modality, i.e., the (non-)existence of multiple optima. These properties are a way of describing fitness landscapes and are often used as a basis for determining suitable algorithms for black-box functions. Here, we propose to train a classifier for such properties using ELA features %(see Section~\ref{sec:ela}) 
as input. That is, we draw a small sample of points in the optimisation problem's search space (e.g., using a Latin hypercube design or random uniform sampling) and compute the corresponding fitness values \cite{Renau2019,kerschke2016_budget}. Subsequently, the fitness landscape of the particular optimisation problem is characterised by means of various numerical summary statistics -- called ELA features \cite{kerschke2019flacco,munoz2015_as,malan2013} -- that are computed based on the sampled points. %Exemplary features are (i) the coefficient of determination $R^2$ of a linear (or quadratic) model trained on the sampled data \cite{mersmann2011}, or (ii) the ratio between the averages distances of the samples' points to their nearest neighbours and nearest better neighbours, respectively \cite{kerschke2015}.

% %% --------------------------------------------------------------------75
\subsection{Method Description}
\label{sec:heatMaps:method}

Our method, outlined in Figure~\ref{fig:scheme2}, is based on previous work by \cite{mersmann2011}, in which various high-level properties of problem landscapes, such as (their degree of) multi-modality and global structure, were discussed. \new{The authors further} (manually) labelled the 24 artificial test functions from the well-known Black-Box Optimization Benchmark (BBOB) \cite{Hansen2009} with \new{the defined} high-level properties.

We use this data -- i.e., the levels of the high-level properties (e.g., \textit{none}, \textit{low}, \textit{medium} or \textit{high} degree of multimodality) -- as outcome or class labels when training classifiers for a total of eight\footnote{As all but one BBOB problem were said to be without plateaus, we removed this property, and instead considered the funnel characteristic from \cite{kerschke2015}.} high-level properties.  %\tea{Where does the number eight come from? We mentioned seven before, shouldn't it be six?}. 
As input, each classification model (one per high-level property) takes a large set of cheap but informative ELA features. For our experiments we considered all 97 features -- like the coefficient of determination $R^2$ that was mentioned in Section~\ref{sec:ela} as an exemplary ELA feature -- from the following nine feature sets: dispersion, level set, meta model, $y$-distribution, angle, information content, nearest better clustering, basic, and principal component analysis \cite{lunacek2006,mersmann2011,kerschke2014,munoz2015ic,kerschke2015}. A detailed, yet compact description of the considered feature sets can be found in the most-recent survey on ELA \cite{kerschke2019flacco}. 

Our experiments are based on all 24 test problems from BBOB and rely on samples of 500 points that were created by means of a latin hypercube design (on the corresponding 10-dimensional search space $[-5, 5]^{10}$). For each of the problems, we then computed the aforementioned 97 ELA features using the R-package \texttt{flacco} \cite{hanster2017,kerschke2019flacco}. In order to capture the general characteristics of a BBOB problem and not traits that are specific to one of its instances\footnote{Each instance is a translated, rotated and/or shifted version of its original function.}, we considered (the first) five problem instances per problem.

Once we had generated appropriately labelled data, we trained a variety of classification models (classification trees \cite{rpart}, random forests \cite{randomforest}, support vector machines \cite{karatzoglou2004} and gradient boosting models \cite{xgboost}) -- all of which have proven to be suitable candidates in the context of feature-based studies (see, e.g., \cite{kerschke2018bbob,kerschke2018tsp}) -- to find strong classifiers for each of the different high-level properties. 
To reduce noise as well as redundancy among the features and thereby improve the quality of the models, each classifier was trained using different greedy (floating forward-backward and backward-forward selection, respectively) and stochastic (evolutionary algorithms with plus-strategy, population size $\mu = 10$, and $\lambda \in \{5, 50\}$ offspring) automated feature selection strategies as presented in \cite{kerschke2018bbob}. 
All models were evaluated using leave-one-function-out cross-validation. 
This allows a fair comparison of the models and reduces the risk of overfitting.

In the end, we obtained \new{at least }one well-performing classification model per high-level property, capable of predicting the respective attribute(s) based on a set of ELA features as input. 
These classifiers can be applied to any black-box function, as long as an appropriate number of samples can be computed. Despite performing well, model predictions are no guarantees and rely on several assumptions. Still, they can serve as sophisticated and computationally efficient guesses for high-level properties of black-box functions.

% %% --------------------------------------------------------------------75
\subsection{Purpose in Context of PCG}

As most search-based PCG approaches employ complex genotype-phenotype mappings as well as simulation-based fitness functions, they must usually be considered as black boxes. In order to make informed decisions about the choice of the different PCG components (representation, fitness function, optimisation algorithm), it is thus crucial to obtain some information on the high-level properties of the resulting fitness landscapes. For example, if the high-level properties indicate a highly multi-modal landscape, restarts of the evolutionary algorithm should be considered.

While diagonal walks (see Section~\ref{sec:walks}) provide a basic way of approaching this lack of information, they can only offer insights into small slices of the complete landscape. Depending on how heterogeneous the landscape is, the information gathered from diagonal walks cannot be \new{generalised} to the entire landscape. A data-driven method, such as the classifiers proposed in this section, can provide information on whichever area of the landscape is sampled. It can thus be used as a way to investigate hypotheses formulated based on domain knowledge or initial observations.

% %% --------------------------------------------------------------------75
\subsection{MarioGAN Results}
\label{sec:heatmap:gbea}

\subsubsection{Analysis}

\begin{figure}[t]
\centering
\includegraphics[width=\textwidth, trim = 1mm 2mm 1mm 1mm, clip]{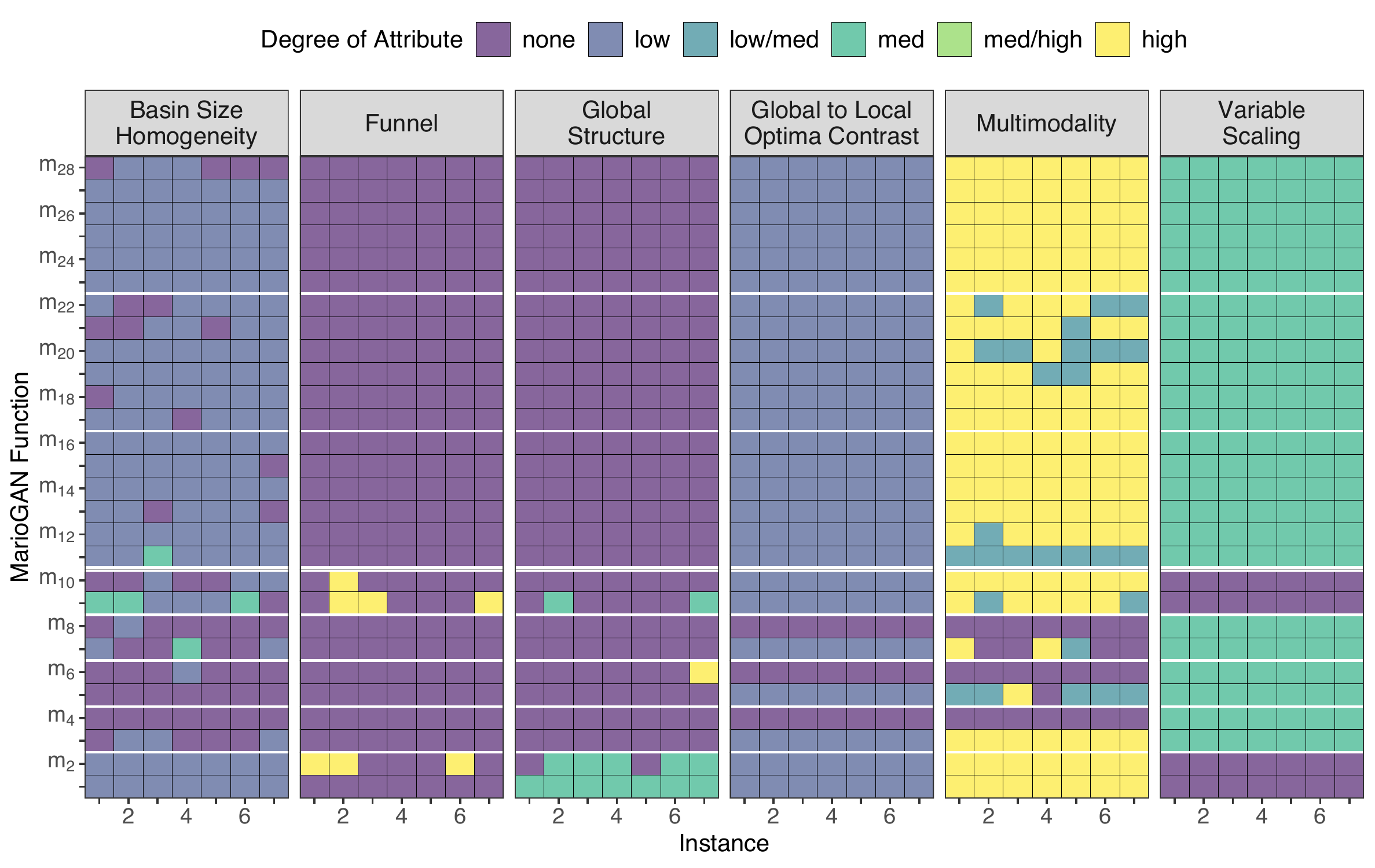}
\caption{Predicted high-level properties for all seven instances of each of the 28 MarioGAN problems. \new{According to our models}, \emph{all} 196 instances are supposedly non-separable and possess highly homogeneous search spaces.}
\label{fig:elaprops}
\end{figure}

The heatmap in Figure~\ref{fig:elaprops} illustrates the predictions for six (of all eight considered) %\tea{Here we say six!} 
high-level properties by previously trained classifiers for all MarioGAN problems. The rows indicate the problems ($m_1$ to $m_{28}$), whereas the columns \new{subdividing each high-level property heading} correspond to the seven instances per problem (see Section~\ref{sec:setup:prob}). 

An immediately noticeable observation is that the high-level properties seem to be relatively similar across all the different MarioGAN problems. Two properties are not even shown because they were constant across all problems: none of the problems are separable (i.e., they cannot be broken down into smaller subproblems), and they all have a very homogeneous search space. 
However, when looking more closely, it is noticeable that for the agent-based problems ($m_{11}$-$m_{28}$) the attributes of the high-level properties are mostly consistent, whereas the characteristics of the representation-based problems display partly considerable differences. For instance, problems based on the fitness measures enemyDistribution ($m_1$, $m_2$) and leniency ($m_9$ and $m_{10}$) show no differences along the different search space dimensions as well as some evidence of a global structure, while all other problems indicate a moderate degree of variable scaling without signs of a funnel or global structure. Another characteristic we observed are the differences between over- and underworld levels for the remaining representation-based problems: The overworld problems ($m_3$, $m_5$ and $m_7$) apparently exhibit a higher degree of multimodality as well as a stronger global to local optima contrast compared to their underground counterparts ($m_4$, $m_6$ and $m_8$).
%
% Summarizing 
\new{Summarising} across all 196 problems,
we seem to be dealing with problems that mostly: 
\begin{compactitem}
    \item[(1)] behave differently in the different dimensions of the search space (\emph{variable scaling}: medium) and are non-separable (\emph{separability}: none), i.e., different dimensions of the search space cannot be treated separately;
    \item[(2)] have no obvious global trends (\emph{funnel}: none, \emph{global structure}: none, \emph{global to local optima contrast}: low) that could be utilised e.g., by estimation of distribution algorithms, such as the Covariance Matrix Adaption Evolution Strategy (CMA-ES) \cite{hansen2006eda};
    \item[(3)] have attraction basins of varying sizes (\emph{basin space homogeneity}: none/low);
    \item[(4)] have a relatively high number of local optima (\emph{multimodality}: high).
\end{compactitem}

Based on these results, we conclude that most of these problems are likely difficult to tackle. This largely aligns with the diagonal walk results discussed in Section~\ref{sec:walks:gbea}.

\subsubsection{Interpretation}

Early results on the optimisation problems indicate that a majority of these problems are indeed difficult to tackle, thus validating the estimations from the classifier. 
For MarioGAN problems, for instance, random search has even shown to be competitive with a wide range of evolutionary algorithms \cite{Volz19}.

This result suggests that the genotype-phenotype mapping used for all problems should be reconsidered, as the problems, independent of fitness function and training set,
have mostly similar high-level properties. However, the goal of PCG approaches is usually not to identify a single best sample of content, but several ideally diverse examples of sufficient quality. Finding a global optimum is thus not the main priority in PCG. The same is also true for multi-modal problems, as well as optimisation in real-world contexts in general.

It should thus be investigated whether the competitive performance of random search can be explained by the abundance of suitable generated levels, which would result in very flat fitness landscapes. In this case, this rich representation should be kept and instead of applying sophisticated algorithms, simple samplers could suffice to identify suitable levels quickly. The same concerns naturally also arise for problems with similar fitness landscapes, which many problems in real-world applications might have. This should thus be a topic for futher investigation.

\subsection{Concluding Remarks}
The approach considered in this section compresses the information of all 97 landscape features into eight high-level properties. %\tea{Here we have eight again}. 
ELA features are intended for data-driven analyses, and the trained models are helpful to gain general insights based on them. In contrast to the previously discussed diagonal walks, the whole search space is considered and thus, global properties can be assigned with reasonable confidence. 
Obviously, the quality of the resulting classification (of the attribute values within each of these properties) strongly depends on (a) the appropriateness of the selected classification model, and (b) the representativeness of the sampled points. Moreover, it should be noted that depending on the complexity of the considered classifier(s), it can be (1) expensive to train a separate classifier (per property), and (2) hard to explain its decisions.

In consequence, the reliance on high-level properties has several issues. They describe complex abstract concepts which require a certain amount of existing familiarity with such properties in order to be interpretable by a user. They also inadvertently shape and thus limit the way in which problems are described. 
The accuracy of the classifier also depends, on the one hand, on the quality of the training data, which must be manually labelled with appropriate high-level properties, and, on the other hand, it is influenced by the similarity between the training data and the data of the test problem(s).

This method is thus suitable only for advanced users. For beginners, other, more intuitive methods are needed to characterise the properties of a given problem.

% %% --------------------------------------------------------------------75
% %% --------------------------------------------------------------------75
\section{Problem Similarity Measures}
\label{sec:tsne}

Another way of characterising problems is to describe them by their similarity to others. Similarity can provide intuition and thus avoid reliance \new{on} complex abstract concepts such as high-level properties discussed in the previous section. In addition, similarity information can also be useful to decide whether findings from different problems are applicable in a new context.

% %% --------------------------------------------------------------------75
\subsection{Method Description}

ELA features (see Section~\ref{sec:heatMaps}) provide a way of characterising and ranking different optimisation problems numerically, and thus constitute a great basis for measuring similarity. They are, however, designed to work as input for data-driven modelling and hence should not be interpreted in isolation. We propose to use dimensionality reduction approaches to capture similarities recorded in ELA features (see Figure~\ref{fig:scheme2}). Here, we specifically consider t-distributed Stochastic Neighbourhood Embedding (t-SNE) \cite{maaten2008} as a tool for finding a low-dimensional, visualisable representation of the problems -- and their (dis)similarities.

As a preparatory step, ELA features are computed for all problems under analysis as described in Section~\ref{sec:heatMaps:method}. This allows a comparison between the problems under investigation. In addition, it is useful to also compute ELA features for other (well-known) problems that can serve as baselines for comparison. Ideally, these problems are well understood and diverse, so that the analysed problems can be characterised by their similarity to the baselines. Good candidates are artificial optimisation problems from benchmarks such as BBOB~\cite{Hansen2009}.

Next, some pre-processing of the data is required. Features with constant values across all problems need to be removed from the data. Given that many features are of different magnitude -- and as we have no intention of interpreting any of them at this point -- all feature vectors are normalised to ensure a more homogeneously scaled data set.

Finally, the dimensionality reduction method t-SNE \cite{maaten2008} is applied to the (normalised) feature data. The resulting representation can be used as a basis for investigating the similarities between problems further.

% %% --------------------------------------------------------------------75
\subsection{Purpose in Context of PCG}

As discussed above, the ability to measure similarity can help in identifying problems that can serve as a suitable (i.e., comparable) test bed for the original problem. This can be helpful for the selection of suitable optimisation algorithms or even for tuning their hyper-parameters. Such a test bed is especially beneficial if it is much cheaper to evaluate than the original problem - which is often quite expensive in simulation-based PCG evaluations.

It might even be possible to identify problems that are similar enough to be able to act as a proxy for the original problem. This could for example be used as a way to identify interesting areas in the search space without having to evaluate the expensive problem.

Similarity measures are \new{also} helpful for comparisons of different versions of the same problem. The effects of modifying the representation within a PCG algorithm, for example, could be tracked this way. Conversely, it can also be used to verify the similarity of the optimisation problems encountered when applying a given PCG approach to different games. If the problems are indeed similar, we can reasonably expect that the \new{problem landscapes exhibit similar characteristics}. In this case, similarity can thus be used as a measure of the robustness of the PCG approach.

Finally, given a sufficient amount of suitable data from different search-based PCG approaches is collected, similarity measures in combination with clustering methods could be used to identify archetypes of PCG problems. This discovery would facilitate the generalisation of PCG approaches and make their outcomes more predictable. This would in turn improve their applicability in industry.

% %% --------------------------------------------------------------------75
\subsection{MarioGAN Results}

\subsubsection{Analysis}

For the analysis, we re-use the ELA features computed for MarioGAN (see Section~\ref{sec:heatmap:gbea}). As baselines, we selected the 10-D problems from the BBOB benchmark \cite{Hansen2009} and a collection of 40 10-D instances that were generated using Shekel's foxhole function\footnote{Eight problems generated with different numbers of peaks (3, 5, 7, 10, 20, 30, 40 and 50). To create five instances each, locations and widths of the peaks were sampled random uniformly from $[0, 10]^{10}$ and $[0, 1]$, respectively.} \cite{shekel}. In previous work, the foxhole function with its large plateaus and small spikes was identified to be potentially similar to a PCG approach based on cellular automata \cite{shekel}. Based on the results from the diagonal walk analysis described in Section~\ref{sec:walks:gbea}, foxhole functions do indeed seem like a good candidate for similarity based on their known landscape characteristics.

\begin{figure}[t]
\centering
\includegraphics[width=0.85\textwidth, trim = 1mm 1mm 1mm 1mm, clip]{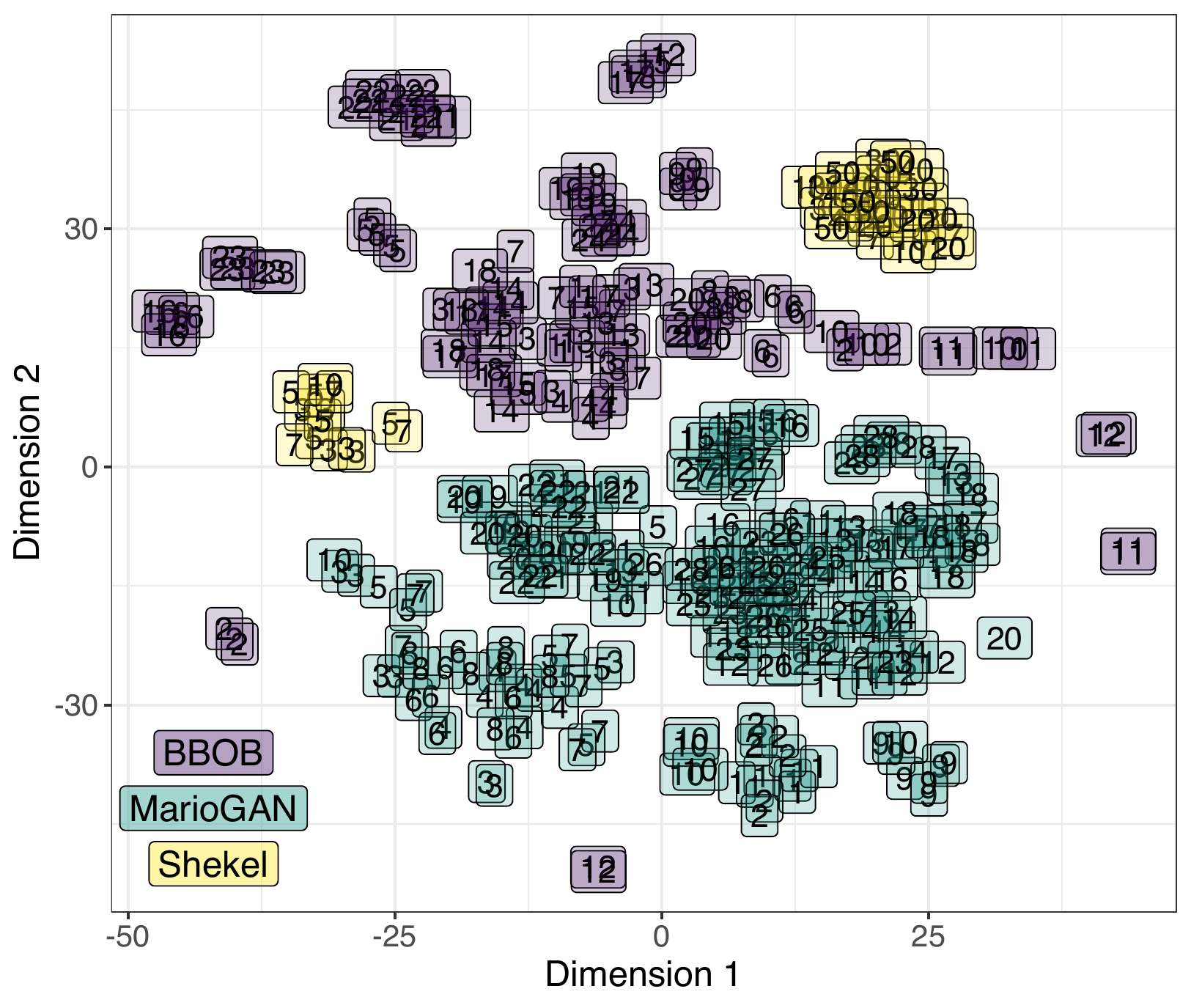}
\caption{Low-dimensional representation of all considered problems from BBOB ($24$ functions $\times$ 5 instances, shown in purple), MarioGAN ($28$ functions (see Table~\ref{mario:over}) $\times~7$ instances, green) and Shekel ($8$ functions $\times~5$ instances, yellow). The mapping was created by applying t-SNE to the corresponding normalised ELA feature vectors.} %\tea{Why is MarioGAN in the legend of a different color than in the plot?}} \bn{The color in the figure is better protected due to overlays, the color in the legend is already bleached out ;)}
\label{fig:ela_tsne}
\end{figure}

Figure~\ref{fig:ela_tsne} provides a two-dimensional representation of the high-dimensional feature data (of all 356 considered problem instances) using t-SNE. 
The problems from MarioGAN (shown in green) form a mostly homogeneous group and do not have any spatial overlap with any of the other benchmark problems. On the other hand, the foxhole functions (yellow) form two groups, which are separated by a large cluster of BBOB problems (purple). 
According to this plot, the three different problem sets are not similar and we thus \new{determine that} the findings from \cite{6463449} \new{do not generalise to MarioGAN}.  

Analysing the MarioGAN problems in more detail, it seems that different instances from the same problem (indicated by the same problem IDs) tend to be similar. It also seems that problems $m_1$-$m_{10}$ with the representation-based fitness functions concentrate on the bottom of the visualisation. They can thus be considered more similar to each other than the agent-based problems. And although this finding \new{was expected, and} could have been observed in the heatmap of high-level properties (see Section~\ref{sec:heatmap:gbea}), the approach considered here is able to show this effect more clearly.

\subsubsection{Interpretation}

This result supports the assumption that the different instances of each problem are actually reasonably similar. It can thus be argued that the results of MarioGAN will be robust to different initialisations of the training process. 

Furthermore, the Mario problems do not resemble any of the artificial problems we compared them against. In consequence, optimisation algorithms that perform well on e.g., BBOB functions won't necessarily perform well in search-based PCG approaches such as Mario level generation.

This finding reinforces the need for more systematic analyses of PCG problems. It might even require specifically suited optimisation algorithms to handle these specific fitness landscapes. A first step towards collecting suitable data for further investigations was made with the proposal of the GBEA \cite{volz2019gbea}, which contains PCG approaches for two different game-based applications. However, in order to allow for more general conclusions, further applications will be needed.

\subsection{Concluding Remarks}
\label{sec:sim:final}

The characterisation of the fitness landscapes determined by dimensionality reduction of the ELA features as proposed here seems to offer a good trade-off between intuitive interpretability and the ability to gain insight into the fitness landscapes on a global level. However, it should be noted that thorough hypothesis testing is required to confirm the findings of the visual \new{analysis}.

We here suggested the usage of t-SNE, as it preserves the local proximity of similar observations instead of just the structure among dissimilar points. This is mainly achieved by minimising the Kullback-Leibler divergence between the (probability mass of the) observations from the high- and low-dimensional spaces \cite{maaten2008}.
Nonetheless, t-SNE is no silver bullet for dimensionality reduction as it also comes with drawbacks: (1) the initial low-dimensional sample created by t-SNE is stochastic and as such, its final result will differ \new{(slightly)} every time, (2) one can not simply add a new instance to an existing plot, instead one has to re-run the entire divergence optimisation procedure, (3) it does not allow for statements regarding the proportion of explained variance (in contrast to classical dimensionality reduction approaches like principal component analysis), and (4) it is computationally much more expensive than its competitors. 

% %% --------------------------------------------------------------------75
% %% --------------------------------------------------------------------75
\section{Summary and Future Work}
\label{sec:final}

In this paper, we demonstrate how tools designed for the systematic analysis of optimisation problems can be employed to understand more about the interaction of the different components (representation, fitness function, search algorithm) of search-based PCG approaches. We further argue how the derived insights can be used to evaluate and improve these three approaches. This discussion is on the one hand executed on a general level, and on the other hand \new{demonstrated with} a level generator for Mario. These results are the basis for our conclusions on the suitability of the different tools we review. Within this study, we focus on analysis tools intended to identify suitable optimisation algorithms. 

We first discussed diagonal walks, which are an easy way of gathering interpretable information on the fitness landscapes in specific areas of the search space. Their disadvantage is their locality and their \new{therefore} very limited perspective regarding global properties of a given optimisation problem.

Next, we investigated a \new{machine learning-based} method that is capable of estimating these high-level properties based on a suitably small number of samples of a PCG problem. Besides potential issues incurred through classification errors, the main downside of this tool is the required familiarity with the high-level properties used to characterise the problems. While this method can thus potentially provide very detailed insights, it is particularly suitable for advanced users. Moreover, feature-based approaches also highly depend on the quality of the sampled points, which includes \new{their amount and distribution}, as they are fundamental for the feature computation -- which in turn \new{forms} the basis for various further investigations.

As a compromise between the very simple diagonal walks and the rather challenging identification of high-level properties, we studied another data-driven -- yet more intuitively interpretable -- method capable of recording an optimisation problem's global properties. We proposed using the similarity to archetypal baseline functions as a way to gain insights into black-box optimisation problems and employed the distance in a learned latent space as distance measure. 

In consequence, the appropriate analysis tool thus needs to be chosen based on the given circumstances and the user's existing expertise. However, as demonstrated in this paper, a simple experiment with diagonal walks can already achieve interesting insights and serve as a baseline and sanity check for any work on search-based procedural content generation. 
Unfortunately, and despite the simplicity of the diagonal walks, even such approaches are rarely performed in related works.

We tested all the tools on an \new{example} application of MarioGAN, a content generator for Mario levels based on generative adversarial networks. Our main finding is that it is likely that the problems will prove difficult to optimise for standard evolutionary approaches. We plan to investigate this further by employing different analysis tools for additional insights from different perspectives in the future. Further, we will validate our assumption by comparing multiple optimisation algorithms on these problems.

Within this paper, we focused on a specific type of analysis and only discussed three selected analysis tools. This was necessary to allow for sufficient depth in our \new{study} and to demonstrate the suitability of the tools on a specific PCG-based use case. However, there exists a wide range of other appropriate tools for systematic analysis of PCG problems with varying degrees of complexity. Examples include the analysis of noise in fitness functions, as well as plots of the Empirical Cumulative Distribution Function (ECDF) for the visualisation of optimisation performance over time~\cite{cocoperf, bartz2020benchmarking}.

In the future, we plan to employ such additional tools to gather more information on PCG problems, starting with the ones that are already part of the game-based benchmark GBEA \cite{volz2019gbea}. Combining observations that were produced with different tools promises further insights into these complex problems. 

Further down the line, we hope to encourage a best practice for making data from different systematic analyses of PCG problems publicly available. Such a dataset would be an invaluable resource in order to conduct research on the robustness and generalisability of search-based PCG approaches. This would allow for a much wider adaptation of such methods in the games industry as the expected performance could then be estimated before implementation. Decision makers in a practical context would thus be able to operate with more information and choose appropriate tools.

\section*{Acknowledgements}

% Boris Naujoks acknowledges the European Commission’s H2020 programme, 
% H2020-MSCA-ITN-2016 UTOPIAE (grant agreement No.\ 722734) as well as the DAAD (German Academic Exchange Service), Project-ID: 57515062 “Multi-objective Optimization for Artificial Intelligence Systems in Industry”. 
%
Tea Tu\v{s}ar acknowledges financial support from the Slovenian Research Agen\-cy (pro\-jects No.\ Z2-8177 and N2-0254, and program No.\ P2-0209). 
Pas\-cal Kersch\-ke acknowledges support by the \href{https://www.ercis.org}{\textit{European Research Center for Information Systems (ERCIS)}}.

\biboptions{sort&compress}
\bibliography{balance-org3} 

\appendix
\section{Function definitions}
\label{sec:appendix}

We formally define here the functions from Section \ref{sec:setup:fitness} that are implemented in the MarioGAN benchmark used for the analysis in the paper. Note that they are all mapped to a minimisation problem via transformation and scaled to fit into a value range of $[0,1]$ in order to better fit into the benchmarking framework which requires strict limits. In many cases, reaching one or both limits of the fitness value range is unrealistic for any meaningful game level.

\begin{compactitem}
    \item[\bf enemyDistribution \cite{marioEval}: ] 
    $1-\dfrac{\sqrt{\dfrac{1}{N}\sum_{i=1}^{N} (x_i - \mu)}}{m_{sd}}$, where $N$ is the number of enemies in the level, $x_i$ is the x-axis coordinate of the i-th enemy, $m_{sd}$ is the largest standard deviation possible given the width of the level and $\mu = \frac{1}{N} \sum_{i=1}^{N} x_i$.  
	\item[\bf positionDistribution \cite{marioEval}: ] 
    $1-\dfrac{\sqrt{\dfrac{1}{N}\sum_{i=1}^{N} (y_i - \mu)}}{m_{sd}}$, where $N$ is the number of tiles in the level that a player can stand on, $y_i$ is the y-axis coordinate of the i-th such tile, $m_{sd}$ is the largest standard deviation possible given the height of the level and $\mu = \frac{1}{N} \sum_{i=1}^{N} y_i$.
	\item[\bf decorationFrequency \cite{marioEval}: ] 
	$1-\dfrac{n_{pt}}{n_{tot}}$, where $n_{pt}$ is the number of \emph{pretty tiles} in the level and $n_{tot}$ the total number of tiles. Pretty tiles are \emph{pretty tiles} := $\{$Tube, Enemy, Destructible Block, Question Mark Block, or Bullet Bill Shooter Column$\}$
	\item[\bf negativeSpace \cite{Canossa2015TowardsAP}: ] 
	$1-\dfrac{n_{st}}{n_{tot}}$, where $n_{st}$ is the number of tiles in the level that the player can stand on and $n_{tot}$ the total number of tiles.
	\item[\bf leniency \cite{mario100}: ] 
	$\frac{1}{2} \left( \frac{v}{n_{tot}}+1 \right)$, where $v = \sum_{i \in P} n_i - \sum_{i \in N} n_i - \dfrac{n_{gaps}}{2} - \overline{d_{gaps}}$ where $P$ contains question blocks with power ups, and $N$ contains bullet bill shooter stations, piranha plant tubes, and all enemy types and $n_i$ is the number of tiles of type $i$. Further, $n_{gaps}$ is the number of gaps the player has to jump over and $\overline{d_{gaps}}$ the average length of all gaps in the level.
	
	The maximum value for $\frac{v}{n_{tot}}$ is $1$ in the unrealistic case that $\sum_{i \in P} n_i =  n_{tot}$ and $\sum_{i \in N} n_i = 0$ and $n_{gaps}=0$, so all tiles in the level are power ups. Conversely, the minimum value for $v$ is $-1$, if $\sum_{i \in N} n_i = n_{tot}$ and $\sum_{i \in P} n_i =  0$, so all tiles are of a type contained in $N$. Gaps can be neglected here because each gap would reduce the computed value by $height$. Therefore, the resulting value of the above formula is between $0$ and $1$.
	\item[\bf basicFitness \cite{togelius2013championship}: ] 
	$\frac{v+0.04}{1.26}$, where $v = (d_{level} - t_{level} + n_{coins} + I_{won}*5000)/5000$, $d_{level}$ is the length of the level passed, $t_{level}$ is the time spent on the level, $n_{coins}$ is the number of gained coins and $I_{won}=1$ if the level was completed and $0$ otherwise. The constants used for normalisation have been determined via experimentation.
	\item[\bf airTime \cite{mariogan}: ] 
	$\begin{cases}
    \frac{t_g}{t_{tot}},& \text{if won}\\
    1,              & \text{otherwise}
\end{cases}$, where $t_g$ is the number of game ticks spent on the ground and $t_{tot}$ the total number of game ticks played.
	\item[\bf timeTaken \cite{mariogan}: ] 
	$\begin{cases}
    1-\frac{t_{tot}}{t_{max}},& \text{if won}\\
    1,              & \text{otherwise}
\end{cases}$, where $t_{tot}$ is the time spent on the level and $t_{max}$ is the total allotted time to finish the level.
\end{compactitem}

\end{document}

%% file: cuboid-draw.tex
\def\tikz@lib@cuboid@setup{%
   \pgfmathsetlengthmacro{\vxx}%
      {\tikz@lib@cuboid@get{xscale}*cos(\tikz@lib@cuboid@get{xangle})*1cm}
   \pgfmathsetlengthmacro{\vxy}%
      {\tikz@lib@cuboid@get{xscale}*sin(\tikz@lib@cuboid@get{xangle})*1cm}
   \pgfmathsetlengthmacro{\vyx}%
      {\tikz@lib@cuboid@get{yscale}*cos(\tikz@lib@cuboid@get{yangle})*1cm}
   \pgfmathsetlengthmacro{\vyy}%
      {\tikz@lib@cuboid@get{yscale}*sin(\tikz@lib@cuboid@get{yangle})*1cm}
   \pgfmathsetlengthmacro{\vzx}%
      {\tikz@lib@cuboid@get{zscale}*cos(\tikz@lib@cuboid@get{zangle})*1cm}
   \pgfmathsetlengthmacro{\vzy}%
      {\tikz@lib@cuboid@get{zscale}*sin(\tikz@lib@cuboid@get{zangle})*1cm}
}

\def\tikz@lib@cuboid@draw#1--#2--#3\pgf@stop{%
    \begin{scope}[join=bevel,x={(\vxx,\vxy)},y={(\vyx,\vyy)},z={(\vzx,\vzy)}]
       \begin{scope}[canvas is yz plane at x=#1]
          \draw[cuboid/all faces,cuboid/edges,cuboid/right face] 
                (0,0) -- ++(#2,0) -- ++(0,-#3) -- ++(-#2,0) -- cycle;
          \draw[cuboid/all grids,cuboid/right grid] (0,0) grid (#2,-#3);
       \end{scope}
       \begin{scope}[canvas is xy plane at z=0]
          \draw[cuboid/all faces,cuboid/edges,cuboid/front face] 
                (0,0) -- ++(#1,0) --  ++(0,#2) -- ++(-#1,0) -- cycle;
          \draw[cuboid/all grids,cuboid/front grid] (0,0) grid (#1,#2);
       \end{scope}
       \begin{scope}[canvas is xz plane at y=#2]
          \draw[cuboid/all faces,cuboid/edges,cuboid/top face] 
                (0,0) -- ++(#1,0) --  ++(0,-#3) -- ++(-#1,0) -- cycle;
          \draw[cuboid/all grids,cuboid/top grid] (0,0) grid (#1,-#3);
       \end{scope}
       \draw[cuboid/hidden edges] (0,#2,-#3) -- (0,0,-#3) -- (0,0,0) 
                (0,0,-#3) -- ++(#1,0,0);
       \begin{scope}[canvas is yz plane at x=#1]
          \draw[cuboid/all faces,cuboid/right face,cuboid/edges,fill opacity=0] 
                (0,0) -- ++(#2,0) -- ++(0,-#3) -- ++(-#2,0) -- cycle;
       \end{scope}
       \begin{scope}[canvas is xy plane at z=0]
          \draw[cuboid/all faces,cuboid/front face,cuboid/edges,fill opacity=0] 
                (0,0) -- ++(#1,0) --  ++(0,#2) -- ++(-#1,0) -- cycle;
       \end{scope}
       \begin{scope}[canvas is xz plane at y=#2]
          \draw[cuboid/all faces,cuboid/top face,cuboid/edges,fill opacity=0] 
                (0,0) -- ++(#1,0) --  ++(0,-#3) -- ++(-#1,0) -- cycle;
       \end{scope}
       \path (0,#2,0) coordinate (-left top front)
                      coordinate (-left front top)
                      coordinate (-top left front)
                      coordinate (-top front left)
                      coordinate (-front top left)
                      coordinate (-front left top);
       \path (0,#2,-#3) coordinate (-left top rear)
                        coordinate (-left rear top)
                        coordinate (-top left rear)
                        coordinate (-top rear left)
                        coordinate (-rear top left)
                        coordinate (-rear left top);
       \path (0,0,-#3) coordinate (-left bottom rear)
                       coordinate (-left rear bottom)
                       coordinate (-bottom left rear)
                       coordinate (-bottom rear left)
                       coordinate (-rear bottom left)
                       coordinate (-rear left bottom);
       \path (0,0,0) coordinate (-left bottom front)
                     coordinate (-left front bottom)
                     coordinate (-bottom left front)
                     coordinate (-bottom front left)
                     coordinate (-front bottom left)
                     coordinate (-front left bottom);
       \path (#1,#2,0) coordinate (-right top front)
                       coordinate (-right front top)
                       coordinate (-top right front)
                       coordinate (-top front right)
                       coordinate (-front top right)
                       coordinate (-front right top);
       \path (#1,#2,-#3) coordinate (-right top rear)
                         coordinate (-right rear top)
                         coordinate (-top right rear)
                         coordinate (-top rear right)
                         coordinate (-rear top right)
                         coordinate (-rear right top);
       \path (#1,0,-#3) coordinate (-right bottom rear)
                        coordinate (-right rear bottom)
                        coordinate (-bottom right rear)
                        coordinate (-bottom rear right)
                        coordinate (-rear bottom right)
                        coordinate (-rear right bottom);
       \path (#1,0,0) coordinate (-right bottom front)
                      coordinate (-right front bottom)
                      coordinate (-bottom right front)
                      coordinate (-bottom front right)
                      coordinate (-front bottom right)
                      coordinate (-front right bottom);
       \coordinate (-left center) at (0,.5*#2,-.5*#3);
       \coordinate (-right center) at (#1,.5*#2,-.5*#3);
       \coordinate (-top center) at (.5*#1,#2,-.5*#3);
       \coordinate (-bottom center) at (.5*#1,0,-.5*#3);
       \coordinate (-front center) at (.5*#1,.5*#2,0);
       \coordinate (-rear center) at (.5*#1,.5*#2,-#3);
       \coordinate (-center) at (.5*#1,.5*#2,-.5*#3);
       \path (0,#2,-.5*#3) coordinate (-left top center) 
                           coordinate (-top left center);
       \path (.5*#1,#2,-#3) coordinate (-top rear center)
                            coordinate (-rear top center);
       \path (#1,#2,-.5*#3) coordinate (-right top center)
                            coordinate (-top right center);
       \path (.5*#1,#2,0) coordinate (-top front center)
                          coordinate (-front top center);
       \path (0,0,-.5*#3) coordinate (-left bottom center) 
                           coordinate (-bottom left center);
       \path (.5*#1,0,-#3) coordinate (-bottom rear center)
                            coordinate (-rear bottom center);
       \path (#1,0,-.5*#3) coordinate (-right bottom center)
                            coordinate (-bottom right center);
       \path (.5*#1,0,0) coordinate (-bottom front center)
                          coordinate (-front bottom center);
       \path (0,.5*#2,0) coordinate (-left front center) 
                           coordinate (-front left center);
       \path (0,.5*#2,-#3) coordinate (-left rear center)
                            coordinate (-rear left center);
       \path (#1,.5*#2,0) coordinate (-right front center)
                            coordinate (-front right center);
       \path (#1,.5*#2,-#3) coordinate (-right rear center)
                          coordinate (-rear right center);
    \end{scope}
}

\tikzset{
  pics/cuboid/.style = {
    setup code = \tikz@lib@cuboid@setup,
    background code = \tikz@lib@cuboid@draw#1\pgf@stop
  },
  pics/cuboid/.default={1--1--1},
  cuboid/.is family,
  cuboid,
  all faces/.style={fill=white},
  all grids/.style={draw=none},
  front face/.style={},
  front grid/.style={},
  right face/.style={},
  right grid/.style={},
  top face/.style={},
  top grid/.style={},
  edges/.style={},
  hidden edges/.style={draw=none},
  xangle/.initial=0,
  yangle/.initial=90,
  zangle/.initial=210,
  xscale/.initial=1,
  yscale/.initial=1,
  zscale/.initial=0.5
}

\newcommand{\tikzcuboidreset}{
\tikzset{cuboid,
  all faces/.style={fill=white},
  all grids/.style={draw=none},
  front face/.style={},
  front grid/.style={},
  right face/.style={},
  right grid/.style={},
  top face/.style={},
  top grid/.style={},
  edges/.style={},
  hidden edges/.style={draw=none},
  xangle=0,
  yangle=90,
  zangle=210,
  xscale=1,
  yscale=1,
  zscale=0.5
}
}

%% file: paper.bbl
\begin{thebibliography}{10}
\expandafter\ifx\csname url\endcsname\relax
  \def\url#1{\texttt{#1}}\fi
\expandafter\ifx\csname urlprefix\endcsname\relax\def\urlprefix{URL }\fi
\expandafter\ifx\csname href\endcsname\relax
  \def\href#1#2{#2} \def\path#1{#1}\fi

\bibitem{pcgbook2}
J.~Togelius, N.~Shaker, The search-based approach, in: N.~Shaker, J.~Togelius,
  M.~J. Nelson (Eds.), Procedural Content Generation in Games: {A} Textbook and
  an Overview of Current Research, Springer, 2016, pp. 17--30.

\bibitem{pcgbook12}
N.~Shaker, G.~Smith, G.~N. Yannakakis, Evaluating content generators, in:
  N.~Shaker, J.~Togelius, M.~J. Nelson (Eds.), Procedural Content Generation in
  Games: {A} Textbook and an Overview of Current Research, Springer, 2016, pp.
  215--224.

\bibitem{kerschke2018survey}
P.~Kerschke, H.~H. Hoos, F.~Neumann, H.~Trautmann, Automated algorithm
  selection: {Survey} and perspectives, Evolutionary Computation 27~(1) (2019)
  3--45.

\bibitem{munoz2015_as}
M.~A. {Mu{\~n}oz Acosta}, Y.~Sun, M.~Kirley, S.~K. Halgamuge, {Algorithm
  Selection for Black-Box Continuous Optimization Problems: A Survey on Methods
  and Challenges}, {Information Sciences (JIS)} 317 (2015) 224~--~245.
\newblock \href {https://doi.org/10.1016/j.ins.2015.05.010}
  {\path{doi:10.1016/j.ins.2015.05.010}}.

\bibitem{kerschke2018bbob}
P.~Kerschke, H.~Trautmann, Automated algorithm selection on continuous
  black-box problems by combining exploratory landscape analysis and machine
  learning, Evolutionary Computation 27~(1) (2019) 99--127.

\bibitem{bischl2012}
B.~Bischl, O.~Mersmann, H.~Trautmann, M.~Preuss, {Algorithm Selection Based on
  Exploratory Landscape Analysis and Cost-Sensitive Learning}, in: {Proceedings
  of the 14th Annual Conference on Genetic and Evolutionary Computation
  (GECCO)}, ACM, 2012, pp. 313~--~320.
\newblock \href {https://doi.org/10.1145/2330163.2330209}
  {\path{doi:10.1145/2330163.2330209}}.

\bibitem{prager2020cc}
R.~P. Prager, H.~Trautmann, H.~Wang, T.~H.~W. B{\"a}ck, P.~Kerschke,
  {Per-Instance Configuration of the Modularized CMA-ES by Means of Classifier
  Chains and Exploratory Landscape Analysis}, {Proceedings of the 2020 IEEE
  Symposium Series on Computational Intelligence (SSCI)} (2020)
  996~--~1003\href {https://doi.org/10.1109/SSCI47803.2020.9308510}
  {\path{doi:10.1109/SSCI47803.2020.9308510}}.

\bibitem{wolpert1997}
D.~H. Wolpert, W.~G. Macready, No free lunch theorems for optimization, IEEE
  Transactions on Evolutionary Computation 1~(1) (1997) 67--82.

\bibitem{malan2013}
K.~M. Malan, A.~P. Engelbrecht, {A Survey of Techniques for Characterising
  Fitness Landscapes and Some Possible Ways Forward}, {Information Sciences
  (JIS)} 241 (2013) 148~--~163.
\newblock \href {https://doi.org/10.1016/j.ins.2013.04.015}
  {\path{doi:10.1016/j.ins.2013.04.015}}.

\bibitem{mersmann2011}
O.~Mersmann, B.~Bischl, H.~Trautmann, M.~Preuss, C.~Weihs, G.~Rudolph,
  Exploratory landscape analysis, in: Genetic and Evolutionary Computation
  Conference (GECCO), ACM, 2011, pp. 829--836.

\bibitem{kerschke2019flacco}
P.~Kerschke, H.~Trautmann, Comprehensive feature-based landscape analysis of
  continuous and constrained optimization problems using the {R}-package
  flacco, in: N.~Bauer, K.~Ickstadt, K.~L{\" u}bke, G.~Szepannek, H.~Trautmann,
  M.~Vichi (Eds.), Applications in Statistical Computing, Springer, 2019, pp.
  93--123.

\bibitem{cocoperf}
N.~Hansen, A.~Auger, D.~Brockhoff, T.~Tu{\v{s}}ar, Anytime performance
  assessment in blackbox optimization benchmarking, IEEE Transactions on
  Evolutionary Computation (2022) 1--1\href
  {https://doi.org/10.1109/TEVC.2022.3210897}
  {\path{doi:10.1109/TEVC.2022.3210897}}.

\bibitem{bartz2020benchmarking}
T.~Bartz-Beielstein, C.~Doerr, D.~{van den Berg}, J.~Bossek, S.~Chandrasekaran,
  T.~Eftimov, A.~Fischbach, P.~Kerschke, W.~La~Cava,
  M.~L{\'o}pez-Ib{\'a}{\~n}ez, K.~M. Malan, J.~H. Moore, B.~Naujoks,
  P.~Orzechowski, V.~Volz, M.~Wagner, T.~Weise, {Benchmarking in Optimization:
  Best Practice and Open Issues}, arXiv preprint arXiv:2007.03488 (2020).

\bibitem{volz2019gbea}
V.~Volz, B.~Naujoks, P.~Kerschke, T.~Tu{\v{s}}ar, Single- and multi-objective
  game-benchmark for evolutionary algorithms, in: Genetic and Evolutionary
  Computation Conference (GECCO), ACM, 2019, pp. 647--655.

\bibitem{munoz2015ic}
M.~A. {Mu{\~{n}}oz Acosta}, M.~Kirley, S.~K. Halgamuge, Exploratory landscape
  analysis of continuous space optimization problems using information content,
  IEEE Transactions on Evolutionary Computation 19~(1) (2015) 74--87.

\bibitem{malan2009}
K.~M. Malan, A.~P. Engelbrecht, Quantifying ruggedness of continuous landscapes
  using entropy, in: IEEE Congress on Evolutionary Computation (CEC), IEEE,
  2009, pp. 1440--1447.

\bibitem{kerschke2015}
P.~Kerschke, M.~Preuss, S.~Wessing, H.~Trautmann, Detecting funnel structures
  by means of exploratory landscape analysis, in: Genetic and Evolutionary
  Computation Conference (GECCO), ACM, 2015, pp. 265--272.

\bibitem{Skvorc2020}
U.~{\v{S}}kvorc, T.~Eftimov, P.~Koro{\v{s}}ec, {Understanding the Problem Space
  in Single-Objective Numerical Optimization Using Exploratory Landscape
  Analysis}, {Applied Soft Computing (ASOC)} 90 (2020) 106138.

\bibitem{munoz2019}
M.~A. {Mu{\~n}oz Acosta}, K.~A. Smith-Miles, {Generating New Space-Filling Test
  Instances for Continuous Black-Box Optimization}, {Evolutionary Computation
  (ECJ)} 28~(3) (2020) 379~--~404.
\newblock \href {https://doi.org/10.1162/evco\_a\_00262}
  {\path{doi:10.1162/evco\_a\_00262}}.

\bibitem{lang2021exploratory}
R.~D. Lang, A.~P. Engelbrecht, {An Exploratory Landscape Analysis-Based
  Benchmark Suite}, Algorithms 14~(3) (2021) 78.

\bibitem{schneider2022hpo}
L.~Schneider, L.~Sch{\"a}permeier, R.~P. Prager, B.~Bischl, H.~Trautmann,
  P.~Kerschke, {HPO $\times$ ELA: Investigating Hyperparameter Optimization
  Landscapes by Means of Exploratory Landscape Analysis}, in: {International
  Conference on Parallel Problem Solving from Nature (PPSN)}, Springer, 2022,
  pp. 575~--~589.

\bibitem{engelbrecht2021influence}
A.~P. Engelbrecht, P.~Bosman, K.~M. Malan, {The Influence of Fitness Landscape
  Characteristics on Particle Swarm Optimisers}, Natural Computing (2021)
  1~--~11.

\bibitem{Pitra2019}
Z.~Pitra, J.~Repick{\`y}, M.~Hole{\v{n}}a, {Landscape Analysis of Gaussian
  Process Surrogates for the Covariance Matrix Adaptation Evolution Strategy},
  in: {Proceedings of the Genetic and Evolutionary Computation Conference
  (GECCO)}, ACM, 2019, pp. 691~--~699.
\newblock \href {https://doi.org/10.1145/3321707.3321861}
  {\path{doi:10.1145/3321707.3321861}}.

\bibitem{Jankovic2020}
A.~Jankovic, C.~Doerr, {Landscape-Aware Fixed-Budget Performance Regression and
  Algorithm Selection for Modular CMA-ES Variants}, in: {Proceedings of the
  2020 Genetic and Evolutionary Computation Conference (GECCO)}, ACM, 2020, pp.
  841~–--~849.
\newblock \href {https://doi.org/10.1145/3377930.3390183}
  {\path{doi:10.1145/3377930.3390183}}.

\bibitem{prager2022automated}
R.~P. Prager, M.~V. Seiler, H.~Trautmann, P.~Kerschke, {Automated Algorithm
  Selection in Single-Objective Continuous Optimization: A Comparative Study of
  Deep Learning and Landscape Analysis Methods}, in: {International Conference
  on Parallel Problem Solving from Nature (PPSN)}, Springer, 2022, pp. 3~--~17.

\bibitem{belkhir2016}
N.~Belkhir, J.~Dr{\'e}o, P.~Sav{\'e}ant, M.~Schoenauer, {Feature Based
  Algorithm Configuration: A Case Study with Differential Evolution}, in:
  J.~Handl, E.~Hart, P.~R. Lewis, M.~L{\'o}pez-Ib{\'a}{\~{n}}ez, G.~Ochoa,
  B.~Paechter (Eds.), {Proceedings of the 14th International Conference on
  Parallel Problem Solving from Nature (PPSN XIV)}, Vol. 9921 of {Lecture Notes
  in Computer Science (LNCS)}, Springer, 2016, pp. 156~--~166.
\newblock \href {https://doi.org/10.1007/978-3-319-45823-6_15}
  {\path{doi:10.1007/978-3-319-45823-6_15}}.

\bibitem{belkhir2017}
N.~Belkhir, J.~Dr{\'{e}}o, P.~Sav{\'{e}}ant, M.~Schoenauer, Per instance
  algorithm configuration of {CMA-ES} with limited budget, in: Genetic and
  Evolutionary Computation Conference (GECCO), ACM, 2017, pp. 681--688.

\bibitem{Renau2019}
Q.~Renau, J.~Dr{\'e}o, C.~Doerr, B.~Doerr, {Expressiveness and Robustness of
  Landscape Features}, in: {Proceedings of the Genetic and Evolutionary
  Computation Conference (GECCO) Companion}, ACM, 2019, pp. 2048~--~2051.
\newblock \href {https://doi.org/10.1145/3319619.3326913}
  {\path{doi:10.1145/3319619.3326913}}.

\bibitem{derbel2019}
B.~Derbel, A.~Liefooghe, S.~Verel, H.~Aguirre, K.~Tanaka, New features for
  continuous exploratory landscape analysis based on the {SOO} tree, in:
  Foundations of Genetic Algorithms (FOGA), ACM, 2019, pp. 72--86.

\bibitem{shaker2016procedural}
N.~Shaker, J.~Togelius, M.~J. Nelson, Procedural Content Generation in Games:
  {A} Textbook and an Overview of Current Research, Springer, 2016.

\bibitem{searchPCG}
J.~Togelius, G.~N. Yannakakis, K.~O. Stanley, C.~Browne, Search-based
  procedural content generation: {A} taxonomy and survery, IEEE Transactions on
  Computational Intelligence and AI in Games 3~(3) (2011) 172--186.

\bibitem{Yannakakis-A}
G.~N. Yannakakis, J.~Togelius, A panorama of artificial and computational
  intelligence in games, IEEE Transactions on Computational Intelligence and AI
  in Games 7~(4) (2015) 317--335.

\bibitem{pcgbook9}
D.~Ashlock, S.~Risi, J.~Togelius, Representations for search-based methods, in:
  N.~Shaker, J.~Togelius, M.~J. Nelson (Eds.), Procedural Content Generation in
  Games: {A} Textbook and an Overview of Current Research, Springer, 2016, pp.
  159--179.

\bibitem{dagstuhl2017}
P.~Spronck, E.~Andr{\'{e}}, M.~Cook, M.~Preu{\ss{}}, Artificial and
  computational intelligence in games: {AI}-driven game design {(Dagstuhl
  Seminar 17471)}, Dagstuhl Reports 7~(11) (2018) 86--129.

\bibitem{Yannakakis-Experience}
G.~N. Yannakakis, J.~Togelius, Experience-driven procedural content generation,
  IEEE Transactions on Affective Computing 2~(3) (2011) 147--161.

\bibitem{mariogan}
V.~Volz, J.~Schrum, J.~Liu, S.~M. Lucas, A.~Smith, S.~Risi, Evolving {Mario}
  levels in the latent space of a deep convolutional generative adversarial
  network, in: Genetic and Evolutionary Computation Conference (GECCO), ACM,
  2018, pp. 221--228.

\bibitem{manifold}
M.~Ganz{\'{a}}lez-Duque, R.~B. Palm, S.~Hauberg, S.~Risi, Mario plays on a
  manifold: Generating functional content in latent space through differential
  geometry, in: 2022 IEEE Conference on Games (CoG), 2022, pp. 385--392.
\newblock \href {https://doi.org/10.1109/CoG51982.2022.9893612}
  {\path{doi:10.1109/CoG51982.2022.9893612}}.

\bibitem{Browne-Automatic}
C.~B. Browne, Automatic generation and evaluation of recombination games, Ph.D.
  thesis, Faculty of Information Technology, Queensland University of
  Technology (2008).

\bibitem{marioEval}
A.~Summerville, J.~R.~H. Mari{\~{n}}o, S.~Snodgrass, S.~Onta{\~{n}}{\'{o}}n,
  L.~Lelis, Understanding {Mario}: {An} evaluation of design metrics for
  platformers, in: Foundations of Digital Games (FDG), ACM, 2017, pp. 1--10.

\bibitem{landscapeautomata}
D.~Ashlock, C.~McGuinness, Landscape automata for search based procedural
  content generation, in: Computational Intelligence in Games (CIG), IEEE,
  2013, pp. 9--16.

\bibitem{preuss2015}
M.~Preuss, Multimodal Optimization by Means of Evolutionary Algorithms,
  Springer, 2015.

\bibitem{hearthstone}
A.~Bhatt, S.~Lee, F.~de~Mesentier~Silva, C.~W. Watson, J.~Togelius, A.~K.
  Hoover, Exploring the hearthstone deck space, in: Foundations of Digital
  Games (FDG), ACM, 2018, pp. 18:1--18:10.

\bibitem{6463449}
D.~Ashlock, S.~McNicholas, Fitness landscapes of evolved apoptotic cellular
  automata, IEEE Transactions on Evolutionary Computation 17~(2) (2013)
  198--212.

\bibitem{Pitzer2012}
E.~Pitzer, M.~Affenzeller, A comprehensive survey on fitness landscape
  analysis, in: J.~Fodor, R.~Klempous, C.~P. {Su{\'{a}}rez Araujo} (Eds.),
  Recent Advances in Intelligent Engineering Systems, Studies in Computational
  Intelligence, Springer, 2012, pp. 161--191.

\bibitem{shekel}
J.~Shekel, Test functions for multimodal search techniques, in: Fifth Annual
  Princeton Conference on Information Science and Systems, 1971, pp. 354--359.

\bibitem{landscapeRL}
S.~Omidshafiei, K.~Tuyls, W.~M. Czarnecki, F.~C. Santos, M.~Rowland, J.~Connor,
  D.~Hennes, P.~Muller, J.~Pérolat, B.~D. Vylder, A.~Gruslys, R.~Munos,
  Navigating the landscape of multiplayer games, Nature Communications
  11~(5603) (2020).

\bibitem{togelius2013championship}
J.~Togelius, N.~Shaker, S.~Karakovskiy, G.~N. Yannakakis, The {Mario AI
  Championship} 2009-2012, AI Magazine 34~(3) (2013) 89--92.

\bibitem{Horn-A}
B.~Horn, S.~Dahlskog, N.~Shaker, G.~Smith, J.~Togelius, A comparative
  evaluation of procedural level generators in the {Mario AI} framework, in:
  Foundations of Digital Games (FDG), Society for the Advancement of the
  Science of Digital Games, 2014, p. 8 pages.

\bibitem{goodfellow2014generative}
I.~Goodfellow, J.~Pouget-Abadie, M.~Mirza, B.~Xu, D.~Warde-Farley, S.~Ozair,
  A.~Courville, Y.~Bengio, Generative adversarial nets, in: Neural Information
  Processing Systems 27 (NIPS), Curran Associates, 2014, pp. 2672--2680.

\bibitem{arjovsky2017wasserstein}
M.~Arjovsky, S.~Chintala, L.~Bottou, Wasserstein generative adversarial
  networks, in: International Conference on Machine Learning (ICML), Vol.~70 of
  Proceedings of Machine Learning Research, PMLR, 2017, pp. 214--223.

\bibitem{Summerville:pcg2016-VGLC}
A.~J. Summerville, S.~Snodgrass, M.~Mateas, S.~O. Villar, The {VGLC:} {The}
  video game level corpus, CoRR abs/1606.07487 (2016).
\newblock \href {http://arxiv.org/abs/1606.07487} {\path{arXiv:1606.07487}}.

\bibitem{Canossa2015TowardsAP}
A.~Canossa, G.~Smith, Towards a procedural evaluation technique: {Metrics} for
  level design, in: Foundations of Digital Games (FDG), Society for the
  Advancement of the Science of Digital Games, 2015, p. 7 pages.

\bibitem{mario100}
N.~Shaker, M.~Nicolau, G.~N. Yannakakis, J.~Togelius, M.~O'Neill, Evolving
  levels for {Super Mario Bros} using grammatical evolution, in: Computational
  Intelligence and Games (CIG), IEEE, 2012, pp. 304--311.

\bibitem{coco}
N.~Hansen, A.~Auger, R.~Ros, O.~Mersmann, T.~Tu{\v{s}}ar, D.~Brockhoff, {COCO:}
  a platform for comparing continuous optimizers in a black-box, Optimization
  Methods Software 36~(1) (2021) 114--144.
\newblock \href {https://doi.org/10.1080/10556788.2020.1808977}
  {\path{doi:10.1080/10556788.2020.1808977}}.

\bibitem{Volz19}
V.~Volz, Uncertainty handling in surrogate assisted optimisation of games, KI
  -- K{\" u}nstliche Intelligenz 34~(1) (2020) 95--99.

\bibitem{rothlauf}
F.~Rothlauf, Representations for Genetic and Evolutionary Algorithms, Vol. 104,
  Springer, 2006, Ch.~2, pp. 73--96.

\bibitem{volzCog2019}
V.~{Volz}, B.~{Naujoks}, On the effects of simulating human decisions in game
  analysis, in: IEEE Conference on Games (CoG), IEEE, 2019, pp. 1--8.

\bibitem{personas}
A.~Liapis, C.~Holmg{\r{a}}rd, G.~N. Yannakakis, J.~Togelius, Procedural
  personas as critics for dungeon generation, in: A.~M. Mora, G.~Squillero
  (Eds.), Applications of Evolutionary Computation, Springer, 2015, pp.
  331--343.

\bibitem{kerschke2016_budget}
P.~Kerschke, M.~Preuss, S.~Wessing, H.~Trautmann, {Low-Budget Exploratory
  Landscape Analysis on Multiple Peaks Models}, in: {Proceedings of the 18th
  Annual Conference on Genetic and Evolutionary Computation (GECCO)}, ACM,
  2016, pp. 229~--~236.
\newblock \href {https://doi.org/10.1145/2908812.2908845}
  {\path{doi:10.1145/2908812.2908845}}.

\bibitem{Hansen2009}
N.~Hansen, S.~Finck, R.~Ros, A.~Auger, Real-parameter black-box optimization
  benchmarking 2009: {Noiseless} functions definitions, Research Report
  RR-6829, {INRIA}, Paris, France (2009).

\bibitem{lunacek2006}
M.~Lunacek, L.~D. Whitley, The dispersion metric and the {CMA} evolution
  strategy, in: Genetic and Evolutionary Computation Conference (GECCO), ACM,
  2006, pp. 477--484.

\bibitem{kerschke2014}
P.~Kerschke, M.~Preuss, C.~I. {Hern\'{a}ndez Castellanos}, O.~Sch{\"{u}}tze,
  J.-Q. Sun, C.~Grimme, G.~Rudolph, B.~Bischl, H.~Trautmann, Cell mapping
  techniques for exploratory landscape analysis, in: EVOLVE -- {A} Bridge
  between Probability, Set Oriented Numerics, and Evolutionary Computation V,
  Springer, 2014, pp. 115--131.

\bibitem{hanster2017}
C.~Hanster, P.~Kerschke, {flaccogui}: {Exploratory} landscape analysis for
  everyone, in: Genetic and Evolutionary Computation Conference (GECCO), ACM,
  2017, pp. 1215--1222.

\bibitem{rpart}
T.~Therneau, B.~Atkinson, B.~Ripley, \texttt{rpart}: {Recursive} Partitioning
  and Regression Trees, {R}-package version 4.1-11 (2017).

\bibitem{randomforest}
A.~Liaw, M.~Wiener, Classification and regression by {RandomForest}, R News
  2~(3) (2002) 18--22.

\bibitem{karatzoglou2004}
A.~Karatzoglou, A.~Smola, K.~Hornik, A.~Zeileis,
  \href{http://www.jstatsoft.org/v11/i09/}{{kernlab -- An S4 Package for Kernel
  Methods in R}}, Journal of Statistical Software (JSS) 11~(9) (2004) 1~--~20.
\newline\urlprefix\url{http://www.jstatsoft.org/v11/i09/}

\bibitem{xgboost}
T.~Chen, T.~He, M.~Benesty, V.~Khotilovich, Y.~Tang, \texttt{xgboost}:
  {Extreme} Gradient Boosting, {R}-package version 0.6-4 (2017).

\bibitem{kerschke2018tsp}
P.~Kerschke, L.~Kotthoff, J.~Bossek, H.~H. Hoos, H.~Trautmann, {Leveraging TSP
  Solver Complementarity through Machine Learning}, {Evolutionary Computation
  Journal (ECJ)} 26 (2018) 597~--~620.
\newblock \href {https://doi.org/10.1162/evco_a_00215}
  {\path{doi:10.1162/evco_a_00215}}.

\bibitem{hansen2006eda}
N.~Hansen, The {CMA} evolution strategy: {A} comparing review, Studies in
  Fuzziness and Soft Computing 192 (2006) 75--102.

\bibitem{maaten2008}
L.~v.~d. Maaten, G.~Hinton, Visualizing data using {t-SNE}, Journal of Machine
  Learning Research 9 (2008) 2579--2605.

\end{thebibliography}
